\documentclass[10pt,twocolumn,letterpaper]{article}

\usepackage{iccv}
\usepackage{times}
\usepackage{epsfig}
\usepackage{graphicx}
\usepackage{amsmath}
\usepackage{amssymb}
\usepackage{csquotes}
\usepackage{soul}
\usepackage{enumerate}
\usepackage[normalem]{ulem}
\usepackage{algorithm}
\usepackage{algorithmicx}
\usepackage{algpseudocode}
\usepackage{booktabs}
\usepackage{array}
\usepackage{rotating}
\usepackage{arydshln}
\usepackage{ragged2e}
\usepackage{multirow}

\usepackage[pagebackref=true,breaklinks=true,colorlinks,bookmarks=false]{hyperref}

\iccvfinalcopy % *** Uncomment this line for the final submission

\def\iccvPaperID{3167} % *** Enter the ICCV Paper ID here

\ificcvfinal\fi

\def\ourmodel{CMINet}
\def\ourdataset{COME15K}

\begin{document}

%%%%%%%%% TITLE
% \title{Cascaded Complementary Learning for RGB-D Saliency Detection}
\title{RGB-D Saliency Detection via Cascaded Mutual Information Minimization}

% \and
\author{
Jing Zhang$^{1}$~
Deng-Ping Fan$^{2,\star}$~
Yuchao Dai$^{3}$~
Xin Yu$^{4}$~ 
Yiran Zhong$^{5}$~ 
Nick Barnes$^{1}$~ 
Ling Shao$^{2}$\\
$^1$ Australian National University\quad
$^2$ IIAI\quad
$^3$ Northwestern Polytechnical University\\
$^4$ University of Technology Sydney\quad
$^5$ SenseTime\\
% \quad
% \and
}

\maketitle
\ificcvfinal\thispagestyle{empty}\fi

%%%%%%%%% ABSTRACT
\begin{abstract}
Existing RGB-D saliency detection models do not explicitly encourage RGB and depth to achieve effective multi-modal learning. In this paper, we introduce a novel multi-stage cascaded learning framework via mutual information minimization to \textbf{explicitly} model the multi-modal information between RGB image and depth data. Specifically, we first map the feature of each mode to a lower dimensional feature vector, and adopt mutual information minimization as a regularizer to reduce the redundancy between appearance features from RGB and geometric features from depth. We then perform multi-stage cascaded learning to impose the mutual information minimization constraint at every stage of the network. Extensive experiments on benchmark RGB-D saliency datasets illustrate the effectiveness of our framework. Further, to prosper the development of this field, 
we contribute the largest (7$\times$ larger than NJU2K) \ourdataset~dataset, which contains 15,625 image pairs with
high quality polygon-/scribble-/object-/instance-/rank-level annotations. Based on these rich labels, we additionally construct four new benchmarks
% \footnote{Code, results, and benchmarks will be made publicly available.} 
with strong baselines and observe some interesting phenomena, which can motivate future model design. Source code and dataset are available at \url{https://github.com/JingZhang617/cascaded_rgbd_sod}.
\end{abstract}

\section{Introduction}
\label{sec:intro}
Saliency detection models are trained to discover the regions of an image that attract human attention. Conventionally, saliency detection is performed mostly
% \sout{merely} \NB{mostly} 
on RGB images only
% \NB{only} 
\cite{BASNet_Sal,SCRN_iccv,F3Net_aaai2020,Iter_Coop_CVPR,A2dele_cvpr2020}. With the availability of
% large amount of 
depth data as shown in Table~\ref{tab:existing_rgbd_dataset},
% \XY{this would give me the impression, there are many depth data and RGBD data are available for SOD.}
% and the provided geometric information, 
RGB-D saliency detection \cite{jing2020uc,dmra_iccv19, zhang2020bilateral,zhao2019Contrast} attracts
% \NB{s} 
great attention. The extra depth data provides real-world geometric information, which is useful for scenarios when the foreground shares similar appearance to
% \NB{to}\sout{as} 
the background. Further, the robustness of depth sensors (\eg Microsoft Kinect) against lighting changes can also benefit the saliency detection task.

\begin{figure}[t!]
   \begin{center}
   \begin{tabular}{ c@{ } c@{ } c@{ } c@{ } c@{ }}
   {\includegraphics[width=0.185\linewidth]{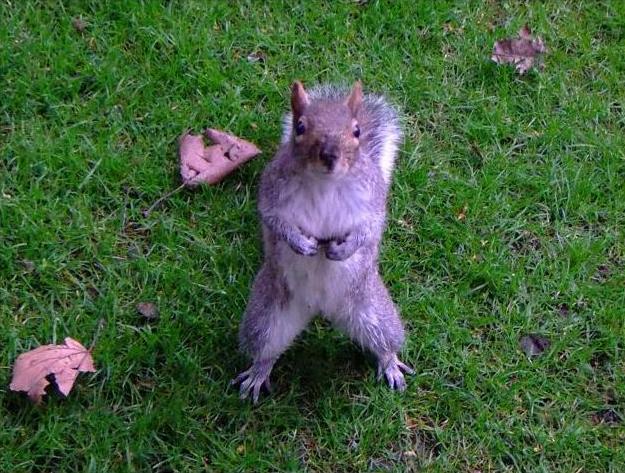}} &
   {\includegraphics[width=0.185\linewidth]{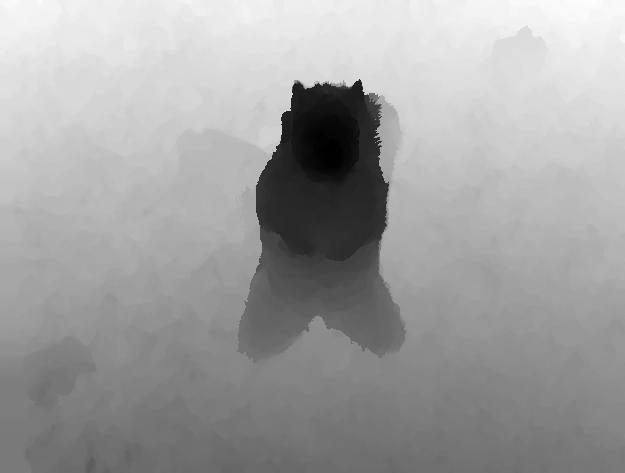}} & {\includegraphics[width=0.185\linewidth]{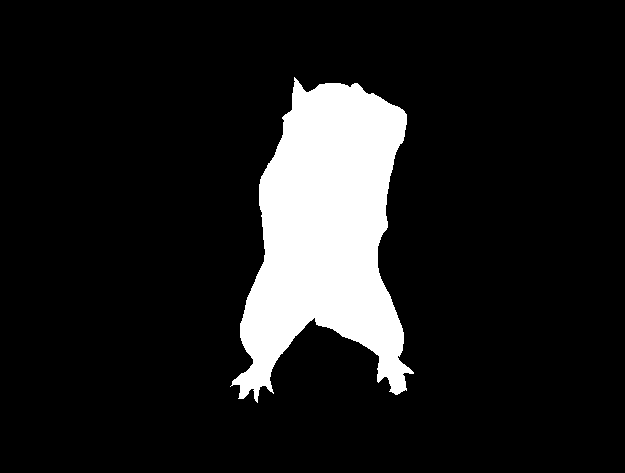}} & {\includegraphics[width=0.185\linewidth]{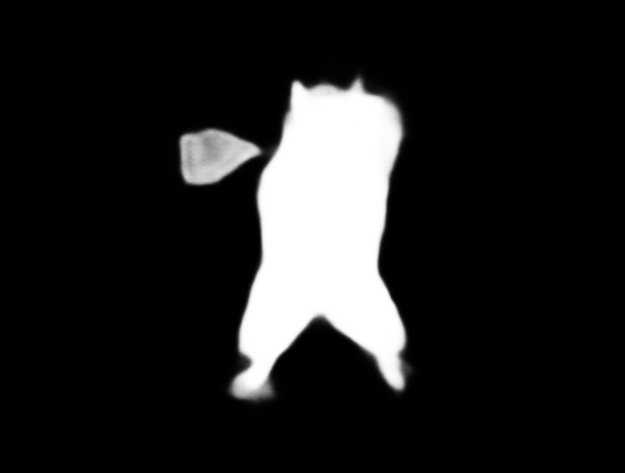}} &
   {\includegraphics[width=0.185\linewidth]{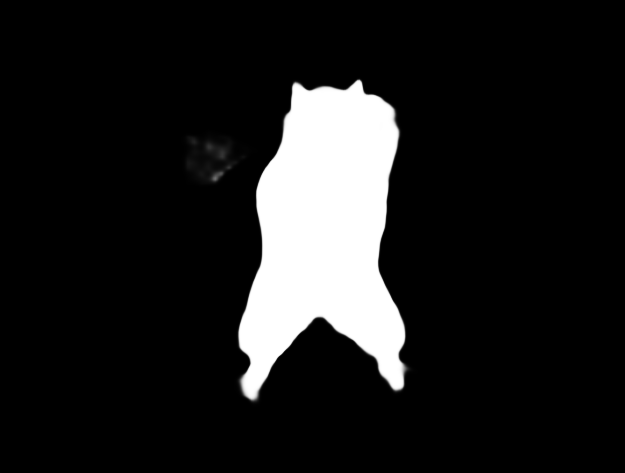}} \\
  {\includegraphics[width=0.185\linewidth]{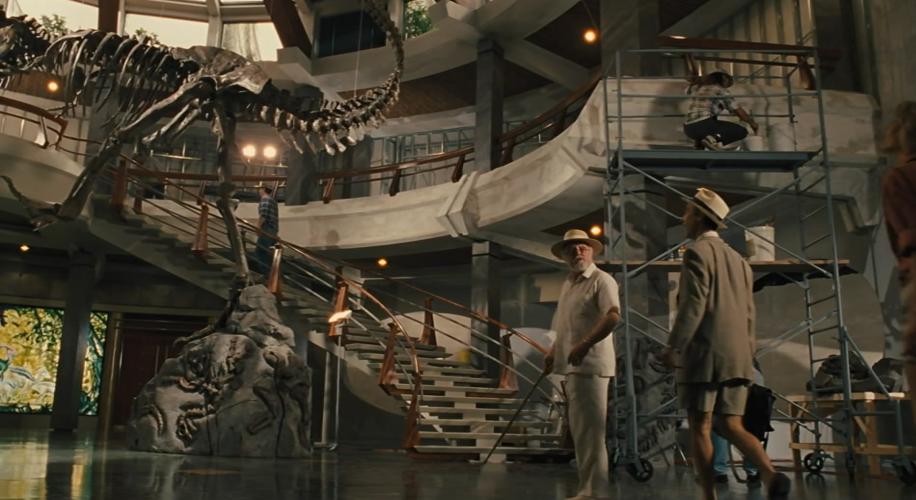}} &
   {\includegraphics[width=0.185\linewidth]{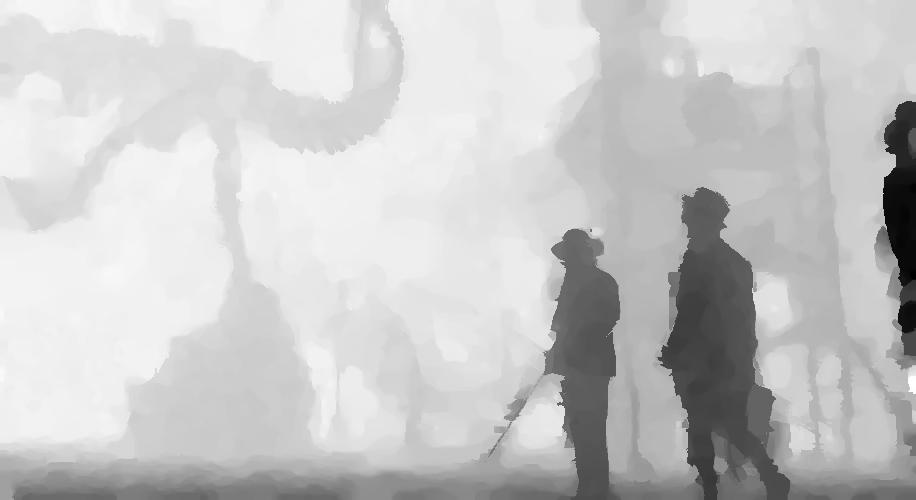}} & {\includegraphics[width=0.185\linewidth]{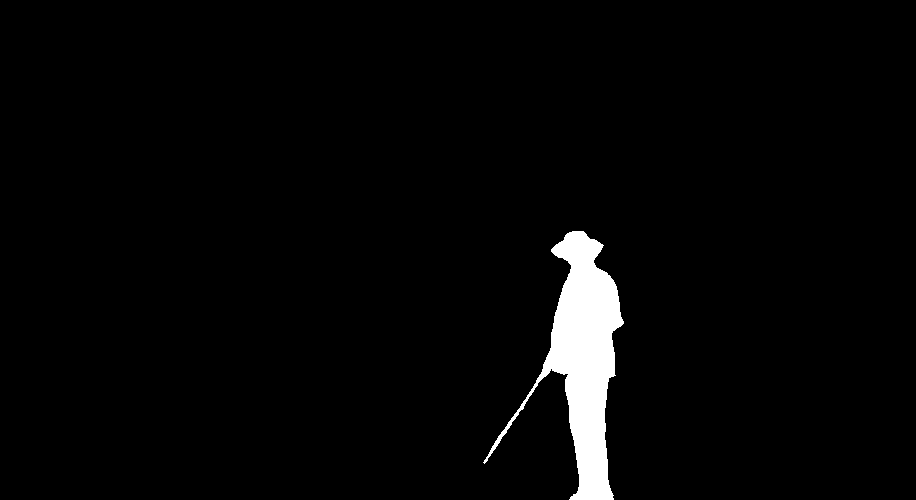}} & {\includegraphics[width=0.185\linewidth]{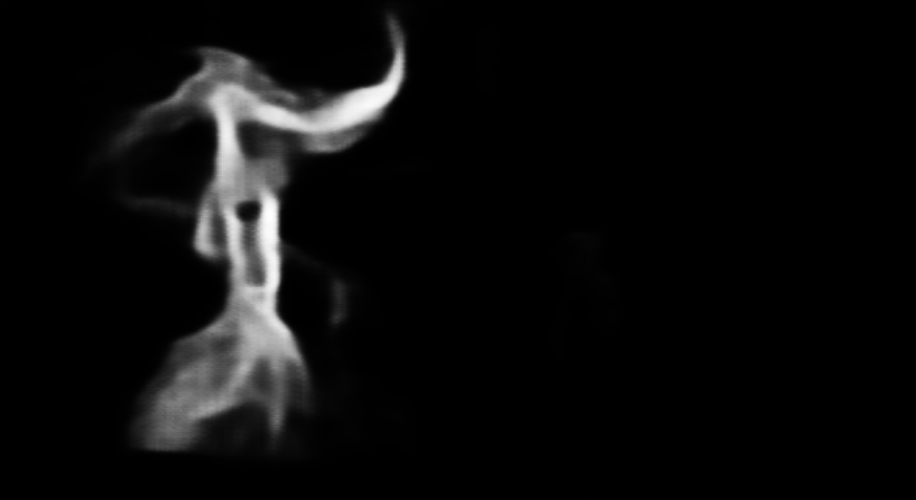}} &
   {\includegraphics[width=0.185\linewidth]{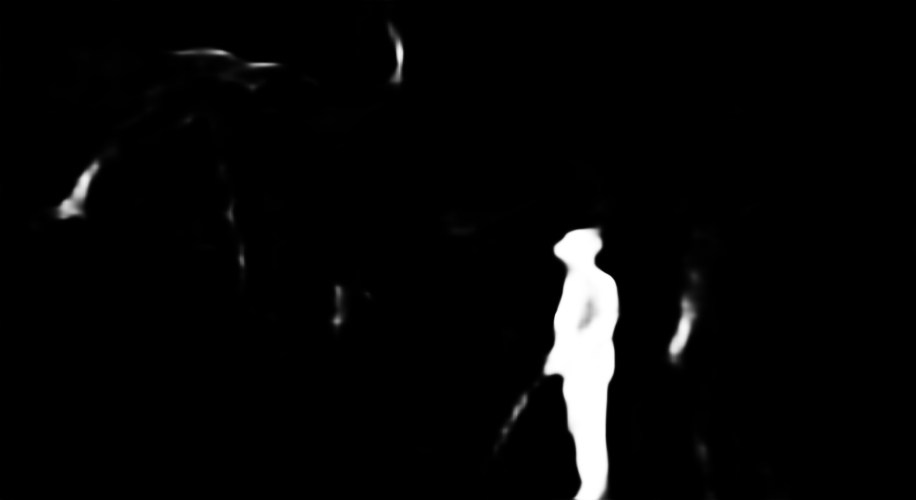}} \\
    \footnotesize{Image} &\footnotesize{Depth} & \footnotesize{GT} & \footnotesize{BBSNet} & \footnotesize{Ours}\\
   \end{tabular}
   \end{center}
   \vspace{-5pt}
   \caption{Comparison of saliency prediction of a state-of-the-art RGB-D saliency detection model, \eg BBSNet \cite{fan2020bbs}, with ours.
%   \XY{Comparison? or just illustration?}
   }
   \label{fig:figure1}
%   \vspace{-4mm}
\end{figure}

As RGB and depth data capture different information about the same scene, existing RGB-D saliency detection models \cite{dmra_iccv19,chen2018progressively,chen2019multi,chen2019three,zhao2019Contrast,ssf_rgbd,self_attention_rgbd,fan2020bbs,ji2020accurate,HDFNet-ECCV2020,zhang2020bilateral} focus on modeling the complementary information of the RGB image and depth data implicitly by using different fusion strategies.
% Towards this goal, existing RGB-D saliency detection models focus on
% % designing networks to 
% fusing the information of these two modes.
Three main fusion strategies have been
% \NB{have been}
%NMB are 
widely studied:
% for RGB-D saliency detection: 
early fusion \cite{qu2017rgbd,jing2020uc}, late fusion \cite{wang2019adaptive,han2017cnns,A2dele_cvpr2020} and cross-level fusion \cite{dmra_iccv19,chen2018progressively,chen2019multi,chen2019three,zhao2019Contrast,ssf_rgbd,self_attention_rgbd,fan2020bbs,ji2020accurate,HDFNet-ECCV2020,zhang2020bilateral,cmms_rgbd}. Although performance improvement can be achieved with effective fusion strategies,
% the three fusion strategies can learn from both RGB and depth data, 
there are
% \NB{are} \sout{exists} 
no constraints on the network design that force
% \NB{
% constraint in the network design to force?} constrain in network designing how to
% force 
it to learn complementary information between
% \NB{from} 
%NB of 
the two modalities, and we cannot explicitly evaluate the contribution of depth data in those models \cite{Zhao2020IsDR}.

\begin{figure*}[tbp]
\centering
% \vspace{-8mm}
   %\begin{left}
   \includegraphics[width=0.86\linewidth]{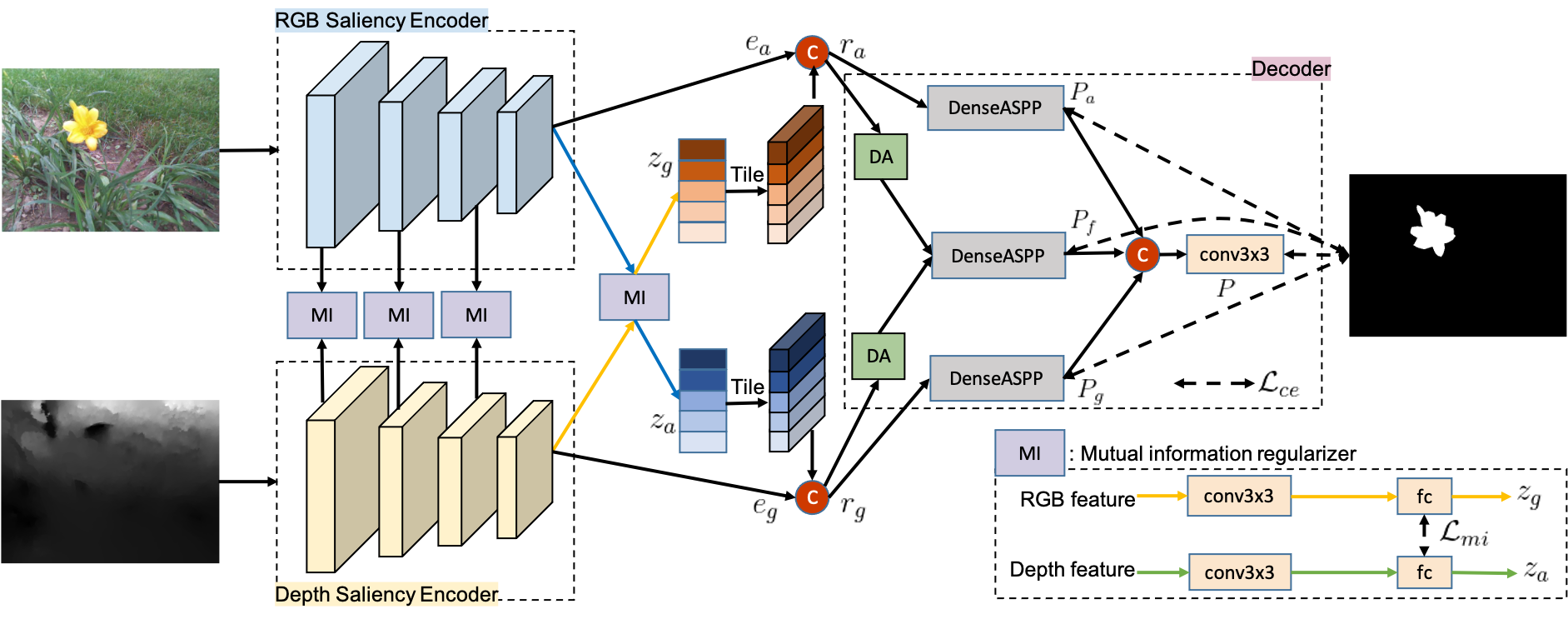}
   %\end{left}
   \vspace{-5pt}
   \caption{Overview of the proposed multi-stage cascaded learning framework for RGB-D saliency detection.
%   Four main modules are included in our framework: 1) a \enquote{RGB Saliency Encoder} module;
% %   for RGB image feature extraction; 
%   2) a \enquote{Depth Saliency Encoder} module;
% %   to extract feature from the depth data; 
%   3) the \enquote{Mutual information regularizer};
% %   to impose the mutual information minimization constraint; 
%   4) a \enquote{Decoder} module to produce our predictions.
   We feed the RGB image and depth to the saliency encoders to extract saliency feature of each mode
%   \sout{modal} \NB{a single one is a mode} 
   with the mutual information regularizer term to push the features to be different from each other. Then, we fuse the lower dimensional feature of each mode ($z_a$ and $z_g$) with raw image feature ($e_a$ and $e_g$) to effectively model the complementary information of each mode and obtain our final prediction $P$. The \enquote{DenseASPP} module is the dense atrous spatial pyramid pooling module from \cite{denseaspp}, and \enquote{DA} is the dual attention module from \cite{fu2019dual}.
%   map both the RGB feature and depth feature to a saliency map in the end in a cross-level fusion way.
   }
   \label{fig:network_overview}
%   \vspace{-4mm}
\end{figure*}

As a multi-modal learning task, a trained model should maximize the joint entropy of different modalities within the network capacity. Maximizing the joint entropy is also equal to the minimization of mutual information, which prevents a network from focusing on redundant information.
To \textit{explicitly} model the
% \NB{the} 
complementary information between
% \NB{between?} 
% %NB of
% \NB{the} 
the RGB image and depth data, we introduce a
% \NB{a?} the 
multi-stage cascaded learning framework
% model based RGB-D saliency detection network
via mutual information minimization.
% \XY{a MIM regularizer. remove as, maybe give it a name? if you are going to use it often}.
% \XY{there are two new contributions, maybe separate it into two sentences.} 
Specifically, we introduce
% \sout{the} 
mutual information minimization as regularizer (as shown in Fig.~\ref{fig:network_overview}) to achieve two main benefits: 1) explicitly modeling the redundancy between
%NB of
% \NB{between} 
appearance features and geometric features;
% with the mutual information minimization regularizer;
2) effectively fusing 
%NB the 
appearance features and geometric features with the mutual information minimization constraint.
% the constraint latent space to achieve multi-mode fusion. 
The produced saliency maps in Fig.~\ref{fig:figure1} illustrate effectiveness of our solution.

Furthermore, we find that
% we observe that
% the existing
% % large-scale 
% RGB image saliency detection training datasets \cite{imagesaliency,msra10k}
% % \NB{s}, 
% % \eg, DUTS \cite{imagesaliency} and MSRA10K \cite{msra10k}, 
% contain more than 10K images,
% % around 10K images for training the RGB saliency detection models, 
% % \XY{maybe unify 10K and 10K? is it 10K = 10,000?}
% %NB the 
% however,
% \NB{however,} \sout{while} 
there is no
% not any 
large-scale RGB-D saliency detection training set. In Table~\ref{tab:existing_rgbd_dataset} we compare the widely used RGB-D saliency datasets,
% \NB{s}, 
in terms of the size,
% of datasets, 
types of data,
% types of the images, 
the sources of depth data, and their roles (for training \enquote{Tr} or for testing \enquote{Te}) in RGB-D saliency detection. 
We note that the conventional training set for RGB-D saliency detection is a
%NB the
combination of samples from the NJU2K \cite{NJU2000} dataset and NLPR \cite{peng2014rgbd}, which includes only 2,200 image pairs in total. Although another 800 training images from the DUT dataset \cite{dmra_iccv19} can serve as the third part of the training set, the total number of training images is 3,000, which is not big enough, and may lead to biased model.
% to produce models of high generalization ability.
% compared with existing RGB saliency detection training sets.
In addition,
%NB Meanwhile, 
we observe that there are
%NB the 
similar backgrounds in the existing RGB-D saliency training set, \eg more than 10\% of the training dataset comes from the same scene with similar illumination conditions.
% same museum with similar light condition. \XY{when you mention museum, you also say it so clearly, I would suggest say come frome the same scene with similar illumination conditions. btw, what is relative related?}
The lack of diversity in the
%NB less diverse 
dataset may render models of poor generalization ability.
% The existing RGB-D saliency detection datasets are not
% % largest training and testing sets have 2K and 1K images respectively, which may not be
% enough to fully evaluate the RGB-D saliency models.
% % in exiting deep learning era. 
% We aim to contribute the largest RGB-D saliency detection dataset
% % (10 times larger) 
% and a new
% % latent variable model based 
% complementary learning framework for explicit multi-mode information fusion.
% which may not be robust to deal with testing images in complex environment.  
% \XY{This paragraph gives me the feeling that only the size of RGBD dataset is smaller. Why small is not good and how does small size affect learning or understanding? Probably providing some comments would make here more convincing. I would believe the large-scale is really necessary. How large would be sufficient.}
% \NB{s}.
% , \eg DUTS training set \cite{imagesaliency} has 10,553 images, and MSRA10K dataset \cite{msra10k} has 10K images.
At the same time, we also notice that the largest testing set \cite{sip_dataset} contains only 1,000 image pairs, which may not be enough to fully evaluate the overall performance of the
%NB those 
deep RGB-D saliency detection models.
% \XY{you can even use one image to evaluate the performance of a model. It may not reflect the statistical characteristics of a model?}

To provide an RGB-D saliency detection dataset for robust model training, and a sufficient size of
% \NB{a sufficient size}
%NB enough 
testing data for model evaluation, we contribute the largest
% (10$\times$scale of previous) 
RGB-D saliency detection dataset, relabeled from Holo50K dataset \cite{hua2020holopix50k}, with 8,025 image pairs for training and 7,600 image pairs for testing. We provide not only binary annotations, but also
% provide an existing RGB-D saliency datasets that often only provide binary ground-truth saliency maps, we also provide
% data\XY{what do you mean ''data'' here?} and
annotations for stereoscopic saliency detection, scribble and polygon annotations for weakly supervised RGB-D saliency detection, instance-level RGB-D saliency annotations and RGB-D saliency ranking.
% For both the
% % \NB{the} 
% training and testing dataset, we provide five types of annotations for both fully supervised learning (binary ground truth saliency maps,
% % \NB{s}\NB{del?: same as exiting RGB-D saliency dataset}, 
% instance RGB-D saliency lablels
% % \NB{do you mean lablels? This is confusing} 
% for instance saliency detection, saliency ranking labels to achieve RGB-D saliency ranking and stereo saliency dataset for stereoscopic saliency detection) and weakly supervised RGB-D saliency detection (scribble annotation and polygon annotation)\footnote{Note most existing RGB-D saliency datasets provide only a binary ground truth map}.
% \NB{perhaps here add: Note most existing RGB-D saliency datasets provide only a binary ground truth map, or put it in a footnote or later} \NB{Alternatively you could say something like: Unlike existing RGB-D saliency datasets that often only provide binary ground truth saliency maps, we also provide data and annotation for RGB-D saliency instance detection, saliency ranking and stereoscopic saliency detection, as well as scribble annotation and polygon annotation for weakly supervised RGB-D saliency detection.}
% Note that, the scribble annotation is our coarse label, and it is refined to achieve pixel-wise accurate annotation as shown in Fig. 
Moreover, we contribute 5,000 unlabeled training images for semi-supervised or self-supervised RGB-D saliency detection.

Our main contributions are: 1)
% summarised as:
We design a multi-stage cascaded learning framework via mutual information minimization for
% based 
RGB-D saliency detection
% network 
to \enquote{explicitly} model redundancy between the
  % \NB{between?} of 
    % \NB{an} 
    RGB image and 
    %NB the 
    depth data.
    2) The mutual information minimization regularizer can be easily extend to other multi-modal learning pipelines to model the redundancy of multiple modalities.
    % \NB{Perhaps something more about specific mechanism?}
    % for effective multi-mode learning.
    % . \NB{This describes it, but do you need to point more tightly to novelty given the review attack?}
    3) We contribute the largest RGB-D saliency detection dataset, with a 15,625 labeled set and a 5,000 unlabeled set to achieve
    % . For the labeled set, we provide five types of annotations for both
    % binary ground truth, instance level annotation, saliency ranking dataset and stereo saliency dataset for 
    fully-/weakly-/un-supervised RGB-D saliency detection.
    % \XY{I think 2 and 3 can be merged together.}
    % We also present scribble and polygon annotation for weakly supervised RGB-D saliency detection.
    4) We present new benchmarks for RGB-D saliency detection, and introduce baseline models for stereoscopic and weakly supervised RGB-D saliency detection.

% \NB{DO you need to mention somewhere in the intro that you are re-labelling an existing dataset?}

%===============================================================
\begin{table}[t!]
  \centering
%   \small
  \scriptsize
  \renewcommand{\arraystretch}{1.2}
  \renewcommand{\tabcolsep}{1.2mm}
  \caption{\small Comparison with the widely used RGB-D datasets.
  }
%   \caption{\small Details of the widely used RGB-D saliency detection dataset, where the \enquote{Type} indicates source of the image, \enquote{Depth Source} represents how the depth data is obtained and \enquote{Role} shows whether the dataset is used for training (Tr) or testing (Te)?
%   }
  \label{tab:existing_rgbd_dataset}
  \begin{tabular}{r|r|c|c|r}
  \hline
    \textbf{Dataset} & \textbf{Size} & \textbf{Type} & \textbf{Depth Source}                     &  \textbf{Role} \\
   \hline
   \hline
    NJU2K\cite{NJU2000}  &1,985  &Movie/Internet  & FujiW3 camera + optical flow  &Tr, Te \\
    DUT \cite{dmra_iccv19} & 1,200  & Indoor/Outdoor & Light-field cameras &Tr, Te \\
    NLPR \cite{peng2014rgbd} &1,000  & Indoor/Outdoor&  Microsoft Kinect &Tr, Te  \\
    SSB \cite{niu2012leveraging}   &1,000  & Internet & Stereo cameras  &Te\\
    SIP \cite{sip_dataset} & 929   & Person in outside & Huawei Mate10 &Te \\
    DES \cite{cheng2014depth}  &135  & Indoor& Microsoft Kinect &Te \\
    LFSD \cite{li2014saliency}   &80  & Indoor/Outdoor& Lytro Illum cameras &Te\\
  \hline
  \textbf{Ours} & \textbf{15,625}   & Indoor/Outdoor & Holopix Social Platform & Tr, Te \\
  \hline

  \end{tabular}
  \vspace{-4mm}
\end{table}

\section{Related Work}

\subsection{RGB-D saliency detection models}
For RGB-D saliency detection, one of the main focuses is to explore the complementary information between the RGB image and the depth data. The former provides appearance information of a scene, while the latter  introduces geometric information. Depending on how information from these two modalities is fused, existing RGB-D saliency detection models can be divided into three categories: early-fusion models \cite{qu2017rgbd,jing2020uc}, late-fusion models \cite{wang2019adaptive,han2017cnns,A2dele_cvpr2020} and cross-level fusion models \cite{dmra_iccv19,chen2018progressively,chen2019multi,chen2019three,zhao2019Contrast,ssf_rgbd,self_attention_rgbd,fan2020bbs,ji2020accurate,HDFNet-ECCV2020,zhang2020bilateral,cmms_rgbd,Li_2020_CMWNet}. 
The first solution directly concatenates the RGB image with its depth,
% information, forming a four-channel input,
while
% , and feed it to the network to obtain both the appearance information and geometric information.
% \noindent\textbf{Early fusion models:} These models directly concatenates the RGB image and the depth data channelwise to obtain a four-channel input data. \cite{DANet,jing2020uc,Fu2020JLDCF}
% \cite{CoNet} define depth as prior, and train an extra depth estimation branch to achieve inference without depth as input.
% \noindent\textbf{Late fusion models:} Instead of fusing in the input layer, t
the late fusion models treat each mode (RGB and depth) separately, and then fusion is achieved at the output layer.
% and two different networks with RGB and depth as inputs are trained to obtain two different predictions for each input RGB-D image pair. Then they fuse the above two predictions at the end of the network.
% \cite{A2dele_cvpr2020}
% \noindent\textbf{Cross-level fusion models:} 
The above two solutions perform multi-modal fusion at the input or output layer, while the cross-level fusion models fuse RGB and depth in the feature space. Specifically, features of an RGB image and depth are gradually fused in different level of the network
\cite{HDFNet-ECCV2020,Li_2020_CMWNet,fan2020bbs,cmms_rgbd,Luo2020CascadeGN,chen2020eccv,ATSA,self_attention_rgbd,ssf_rgbd}.
Although the existing methods fuse the RGB image and depth data for multi-modal learning,
% are used in network learning, 
none of them \textit{explicitly} illustrate
% \XY{models explicitly model. lets change another word.} 
how the network achieve effective multi-modal learning.
% \NB{are you still using this?:complementary information} is learnt. 
We propose a cross-level fusion model as shown in Fig.~\ref{fig:network_overview}. By designing the \enquote{Mutual information regularizer}, we aim to reduce redundancy of appearance features and geometric features for effective multi-modal learning.

\subsection{Multi-modal learning with RGB-D dataset}
% Complementary learning is widely used in multi-mode learning systems for effective information fusion.
% ly fuse information from each mode. 
The basic assumption for multi-modal learning is that there exits both common and diverse information in the separate modalities. For the RGB-D dataset, the RGB image and depth data share similar semantic information, which can be defined as the common parts. The RGB image encodes the appearance information, including intensity or color of the object, while the depth data encodes geometric information, showing the relative geometric localization of the objects. The difference between appearance information and geometric information is the diverse part of these two modalities. The main focus to achieve multi-modal learning for RGB-D data is through using different fusion strategies \cite{wang2018depthconv,real_time_rgbd_semantic,Luo2020CascadeGN,ASIF-Net}, \eg early fusion, late fusion or cross-level fusion. Different from conventional solutions, we introduce a multi-stage cascaded learning framework via mutual information minimization to reduce the feature redundancy of each modal. Although mutual information maximization \cite{Informax_book,TscDjoRubGelLuc20} is widely used in representation learning to produce a representation that is similar to the input, we take the mutual information minimization as a regularizer
% \XY{which one would be better a regularizer or regularization?}
to reduce the feature redundancy for effective multi-modal learning.
% model based complementary learning framework to explicitly model the complementary information in latent space.

\subsection{RGB-D saliency datasets}
The widely used RGB-D saliency detection datasets include NJU2K \cite{NJU2000}, NLPR \cite{peng2014rgbd}, SSB \cite{niu2012leveraging}, DES \cite{cheng2014depth}, LFSD \cite{li2014saliency}, SIP \cite{sip_dataset}, DUT \cite{dmra_iccv19}, \etc, as shown in Table \ref{tab:existing_rgbd_dataset}. The typical training dataset
% RGB-D saliency detection 
is the combination of 1,485 images from NJU2K \cite{NJU2000} and 700 images from NLPR \cite{peng2014rgbd}. Piao \etal~\cite{dmra_iccv19} introduces the DUT dataset, with 800 images
% to the  
for training and 400 images for testing.
% To test model performance on DUT testing dataset, they suggest to further fine-tune the trained model on the DUT training set. 
To prosper the RGB-D saliency detection task, we introduce the largest RGB-D saliency detection training and testing dataset, which will be introduced in Section \ref{the_new_dataset}.
% , which includes 8,025 RGB-D images pairs with high quality ground truth saliency maps for training, and another 7,600 image pairs for testing.\XY{do you think we can remove this? It might be a bit repetitive.}
% As our dataset is large enough, no training set combination or further fine-tune is needed. 

\section{Proposed \ourmodel}
\label{our_method}

We introduce a multi-stage cascaded learning
% based complementary learning 
framework in Fig.~\ref{fig:network_overview} to explicitly model 
%NB the 
complementary information for RGB-D saliency detection.
% Specifically, we first present the \enquote{Saliency Encoder} branch for saliency feature extraction. Then the \enquote{Latent Feature} module is designed to obtain latent feature of each mode, which will be fed to the \enquote{Complementary learning} module to explicitly reduce the information redundancy of RGB image and depth data. Then the \enquote{Saliency Decoder} module is designed to produce our final prediction.

%  Meanwhile, we fuse the appearance information and geometric information with a \enquote{Saliency Decoder} to produce our final saliency prediction $P_i$. 
\subsection{Saliency encoder}
We denote our training dataset as $T=\{X_i,Y_i\}_{i=1}^N$, where $i$ indexes the images and $N$ is the size of the training set, $X_i$ and $Y_i$ are the input RGB-D image pair and its
% \NB{it's is a shortening of it is: it's} 
corresponding ground-truth (GT) saliency map.
We feed the training image pairs (RGB image $I$ and
% \sout{the} 
depth $D$) to the saliency encoder,
% \NB{I think having lots of quotes like this looks a bit excessive - perhaps just name it}
as illustrated in Fig.~\ref{fig:network_overview}, to extract appearance features $f_{\alpha_a}(I)$ and geometric features $f_{\alpha_g}(D)$ respectively, where $\alpha_a$ and $\alpha_g$ are the parameters of our RGB saliency encoder and depth saliency encoder respectively.

We build the saliency encoders upon the ResNet50 network \cite{ResHe2015}, which includes four convolutional stages $\{s^1,s^2,s^3, s^4\}$. We add one additional convolutional layer of kernel size $3\times3$ after each $s^c \in \{s^c\}_{c=1}^4$ to reduce the channel dimension of $s^c$ to $C=32$, and obtain feature maps $\{e^1,e^2,e^3,e^4\}$.
% Then we adopt one DenseASPP \cite{denseaspp} layer after $e^4$ to obtain saliency map $S_a$ (or $S_g$ for the depth saliency encoder branch) for the RGB image based saliency encoder branch. 
The final output of the RGB saliency encoder module is $e_a=\{e_a^1,e_a^2,e_a^3,e_a^4\}$, and that of the depth saliency encoder is $e_g=\{e_g^1,e_g^2,e_g^3,e_g^4\}$. Note that, the RGB saliency encoder and depth saliency encoder share the same network structure but not weights.

\subsection{Feature embedding}
\label{latent_feature_sec}
% Then we map the above two different features to the \enquote{Latent Feature} network to obtain latent feature $z_a$ and $z_g$ representing appearance and geometric feature respectively. We use the \enquote{Mutual Information Minimization} as a regularizer to explicitly force the the two features sharing limited information for complementary information exploring.

% Different from existing RGB-D saliency detection models which implicitly model the complementary information of RGB image and depth data through carefully designed fusion strategy, we introduce the first latent variable model based complementary learning pipeline to explicitly model the complementary information of each mode, as shown in Fig. \ref{fig:network_overview}. 
Given the output $e_a = \{e_a^1,e_a^2,e_a^3,e_a^4\}$ from the RGB saliency encoder and $e_g = \{e_g^1,e_g^2,e_g^3,e_g^4\}$ from the depth saliency encoder, we aim to map both the RGB feature and depth feature to a lower-dimensional feature space for feature embedding.
% , which will to map $e_a$ and $e_g$ to
% % for the depth data is obtained in the same way.) training dataset $T=\{X,Y\}$, where $X=\{I,D\}$ is the image pair of RGB image $I$ and depth data $D$. We have already obtain their saliency feature $e_r$ and $e_d$. Then we introduce another variational encoder network $f_r$ and $f_d$ to map $e_r$ and $e_d$ to 
% latent feature $z_a = f_{\beta_a}(e_a)$ and $z_g = f_{\beta_g}(e_g)$ respectively, where $\beta_a$ and $\beta_g$ are the parameters of the latent feature module for RGB images and depth respectively.
% 
Specifically, we propose a multi-stage cascaded learning strategy to perform the complementary learning at each stage of the network. For the lower stages, we feed the RGB feature $\{e_a^c\}_{c=1}^3$ and the depth feature $\{e_g^c\}_{c=1}^3$ to two different $3\times3$ convolutional layers (\enquote{conv3x3} in Fig.~\ref{fig:network_overview}) to obtain feature maps of channel size $4*C$ for both the RGB branch and depth branch. Then we adopt two fully connected layers (\enquote{fc} in Fig.~\ref{fig:network_overview}) to map the feature map of channel size $4*C$ to two different lower-dimensional feature vectors $\{z_a^c\}_{c=1}^3$ and $\{z_g^c\}_{c=1}^3$ of size $K=6$ respectively. The complementary learning related loss (which will be introduced in Section \ref{sub_sec_complementary} and \ref{obj_fun_sec}) is adopted to reduce the feature redundancy of RGB and depth at lower stages. 
At the highest stage, we first tile the lower dimensional feature vector $z_a^4$ and $z_g^4$ in spatial dimension. Then, we concatenate them with raw image feature\footnote{We define the concatenation of $\{e_a^c\}_{c=1}^4$ as the raw RGB feature and the concatenation of $\{e_g^c\}_{c=1}^4$ as raw depth feature} of the other mode
% .  are tiled and concatenated with the depth saliency feature (the concatenation of $\{e_g^c\}_{c=1}^4$) and RGB saliency feature (the concatenation of $\{e_a^c\}_{c=1}^4$) respectively, 
to obtain the $4*C+K$ channel size feature map $r_a$ and $r_g$ for the RGB branch and depth branch respectively.

\begin{figure*}[t!]
   \begin{center}
   \begin{tabular}{ c@{ } c@{ } c@{ } c@{ } c@{ } c@{ } c@{ }}
   {\includegraphics[width=0.13\linewidth]{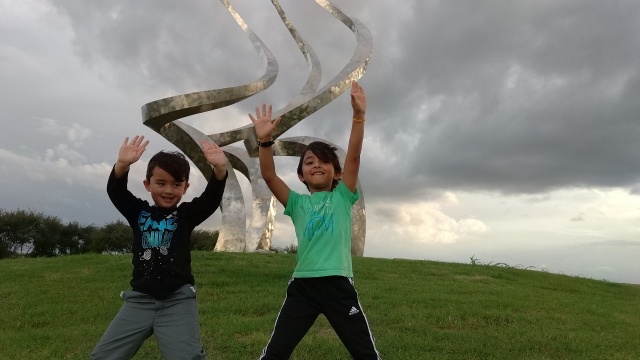}} 
   & {\includegraphics[width=0.13\linewidth]{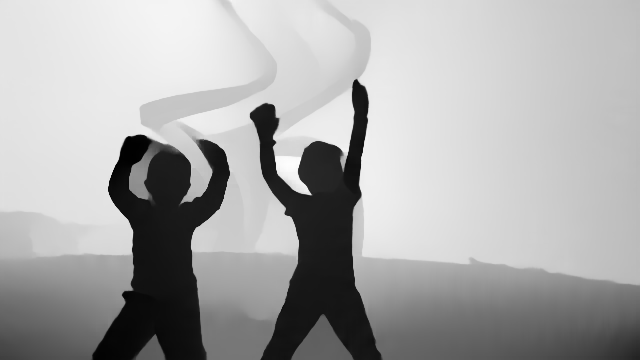}} 
   & {\includegraphics[width=0.13\linewidth]{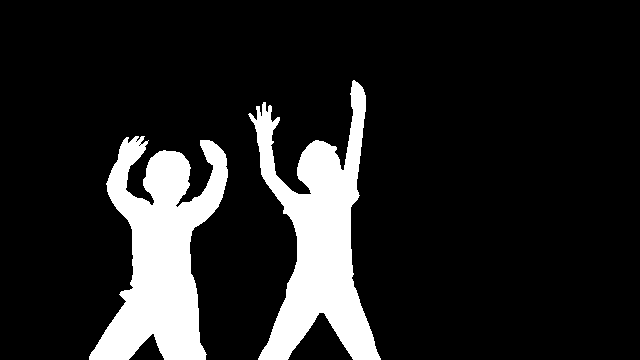}} 
   & {\includegraphics[width=0.13\linewidth]{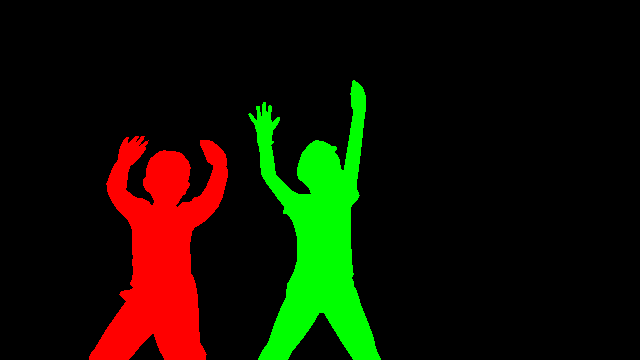}}
   & {\includegraphics[width=0.13\linewidth]{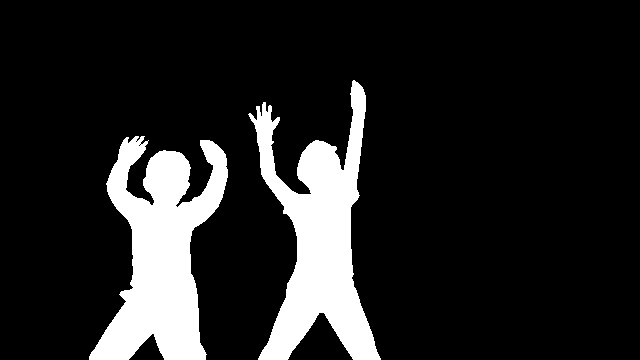}}
   & {\includegraphics[width=0.13\linewidth]{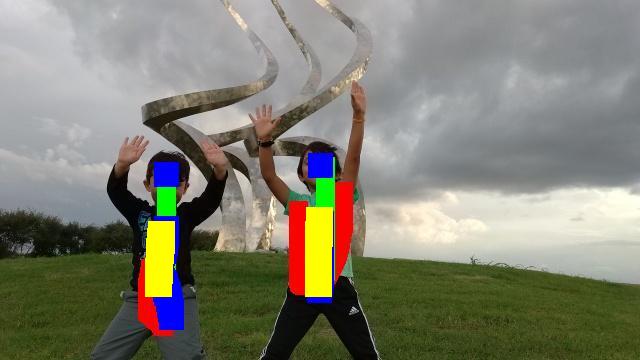}}
   & {\includegraphics[width=0.13\linewidth]{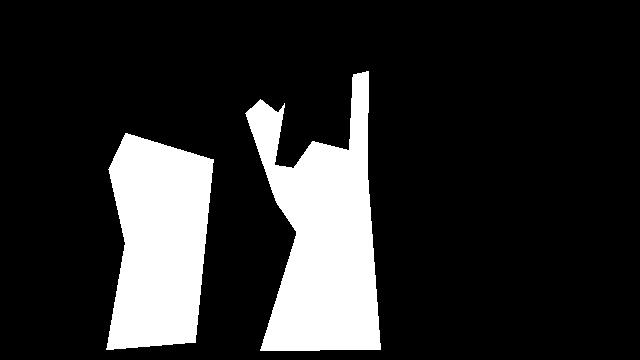}} \\
   {\includegraphics[width=0.13\linewidth]{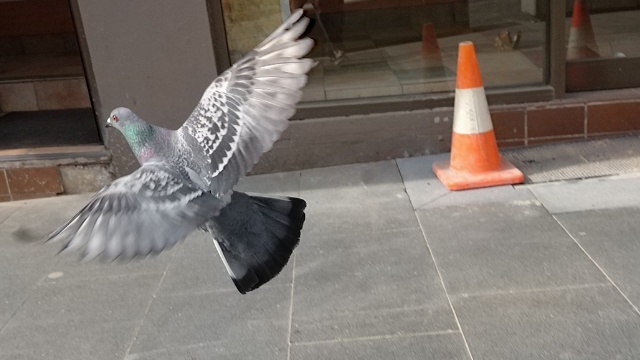}} 
   & {\includegraphics[width=0.13\linewidth]{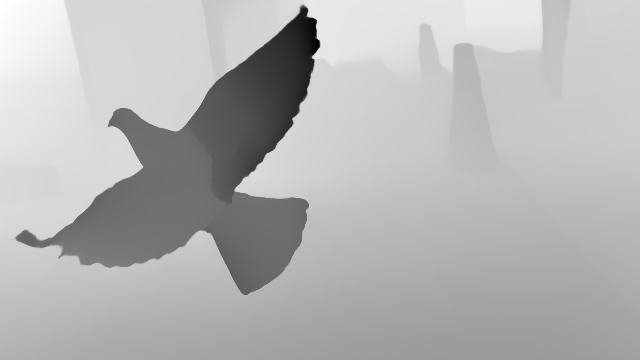}} 
   & {\includegraphics[width=0.13\linewidth]{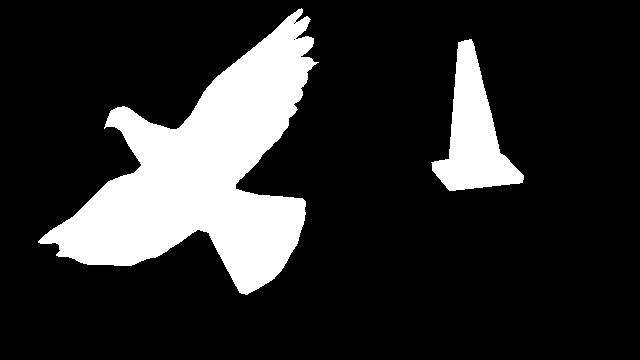}} 
   & {\includegraphics[width=0.13\linewidth]{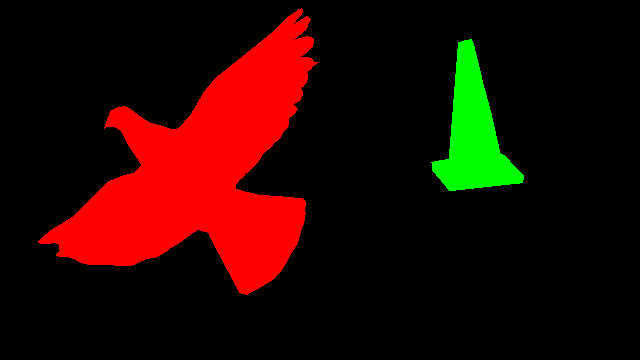}}
   & {\includegraphics[width=0.13\linewidth]{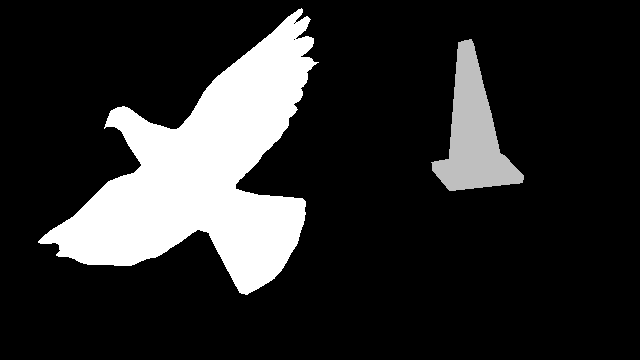}}
   & {\includegraphics[width=0.13\linewidth]{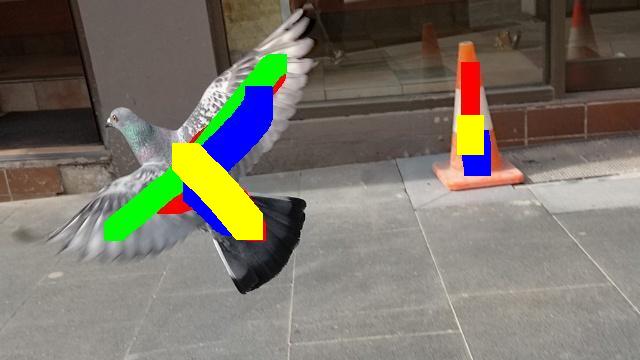}}
   & {\includegraphics[width=0.13\linewidth]{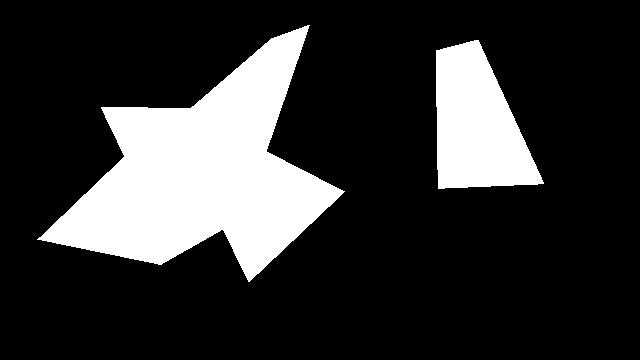}} \\
    \footnotesize{(a)} &\footnotesize{(b)} & \footnotesize{(c)} & \footnotesize{(d)} &\footnotesize{(e)} & \footnotesize{(f)} & \footnotesize{(g)}\\
   \end{tabular}
   \end{center}
   \vspace{-5pt}
   \caption{Annotations of our new RGB-D saliency detection datasets: (a) the RGB image, (b) depth data and (c) binary ground truth,
%   . We provide extra annotations:
   (d) instance level annotation, (e) ranking based annotation, (f) scribble annotation and (g) polygon annotation. Our diverse annotations will facilitate developing different fully/weakly supervised RGB-D saliency detection.}
   \label{fig:dataset_annotation_all}
%   \vspace{-4mm}
\end{figure*}

\subsection{Multi-modal learning}
\label{sub_sec_complementary}
After obtaining the feature embedding $z_a$ and $z_g$ for the
% \NB{the} 
RGB image and depth data, we introduce a
% \NB{a} 
mutual information minimization regularizer to explicitly reduce the redundancy between
% \NB{between?} of 
these two modalities. Our basic assumption is that a good appearance saliency feature and geometric saliency feature pair should carry both common parts (semantic related) and different attributes (domain related). 
%NB The m
Mutual information $M_I$ is used to measure the difference between
% \NB{between} 
%NB of 
the entropy terms:
% \begin{equation}
% \label{mutual_information}
%     I(z_a,z_g) = H(z_a)-H(z_a|z_g)=H(z_g)-H(z_g|z_a),
% \end{equation}
\begin{equation}
\label{mutual_information}
    M_I(z_a,z_g) = H(z_a) + H(z_g) - H(z_a, z_g),
\end{equation}
where $H(.)$ is the entropy, $H(z_a)$ and $H(z_g)$ are marginal entropies, and $H(z_a, z_g)$ is the joint entropy of $z_a$ and $z_g$. Intuitively, we have the Kullback–Leibler divergence (KL) of the two latent variable (or the conditional entropies) as:
\begin{equation}
\label{symetic_kl1}
    KL(z_a||z_g) = H_{z_g}(z_a) - H(z_a),
\end{equation}
\begin{equation}
\label{symetic_kl2}
    KL(z_g||z_a) = H_{z_a}(z_g) - H(z_g),
\end{equation}
where $H_{z_g}(z_a)=-\sum_x z_a(x)\log z_g(x)$ is the cross-entropy.
We then sum Eq.~\ref{mutual_information}, Eq.~\ref{symetic_kl1} and Eq.~\ref{symetic_kl2}, and obtain:
\begin{equation}
\label{mutual_infor_kl}
\begin{aligned}
     M_I(z_a,z_g) = & H_{z_g}(z_a) + H_{z_a}(z_g) - \\
    & H(z_a, z_g) - (KL(z_a||z_g) + KL(z_g||z_a)).
\end{aligned}
\end{equation}

Given the RGB image and depth data, $H(z_a, z_g)$ is non-negative, then
%NB . Then, 
minimizing the mutual information can be achieved by minimizing:
% the negative symmetric KL term:
$\mathcal{L}_{mi}=(H_{z_g}(z_a) + H_{z_a}(z_g))-(KL(z_a||z_g) + KL(z_g||z_a))$.
% Eq.~\eqref{mutual_infor_kl} further suggests that mutual information is a good measure for dependency or complementary.
% \NB{complementarity?} complementary. 
Intuitively,
% \XY{maximizing M is to reduce} 
$M_I(z_a,z_g)$ measures the reduction of uncertainty in $z_a$ when $z_g$ is observed, or vice versa. As a multi-modal learning task, each mode should learn some new attributes of the task from other modalities. By minimizing $M_I(z_a,z_g)$, we can effectively explore the complementary attributes of both modalities.
% , thus reduce the modal redundancy.
Note that, although KL loss term was used in \cite{jing2020uc} as distribution similarity measure,
% to measure the similarity of the distribution of the latent variable. W
we use it to measure mode similarity for multi-modal learning.
% to encourage the difference between
% % \NB{between} 
% %NB of 
% the two modes.
% While, as the two mode are designed for the same task, we still hope that they have some common parts.
% To explicitly model the complementary information of each mode, we introduce mutual information minimization constrain similar to InfoGAN \cite{infogan} to measure the similarity of appearance information and geometric information in the latent space.

% Further,
%NB Meanwhile, 
% As $z_a$ and $z_g$ encode the appearance and geometric information respectively, we intend to fuse the above features in the feature space to achieve effective multi-mode
% % \NB{this is generally known as multi-mode fusion - you would have to change it throughout if you change it} 
% fusion. Specifically, we map $e_a^4$ from the RGB saliency encoder branch to a $K=32$ dimensional feature vector by using one fully connected layer. Then we concatenate it with $z_g$, and map the concatenated feature with one DenseASPP~\cite{denseaspp} to obtain the RGB saliency prediction $P_a$. Similarly, we can obtain the depth saliency prediction $P_g$ by fusing $e_g^4$ with $z_a$.
% % \XY{the RGB and depth saliency predictions are supervised by GT saliency. Right? In that case, need to add one or two sentences why concatenating ea4 with za to achieve RGB sal.}

\subsection{Saliency decoder}
With the mutual information as a regularizer, we achieve the feature redundancy constraint at the lower stages of the network, and obtain the refined RGB saliency feature $r_a$ and refined depth saliency feature $r_g$ at the highest stage. We then adopt one DenseASPP~\cite{denseaspp} module after $r_a$ to obtain the RGB saliency prediction $P_a$ with multi-scale context information. Similarly, we can obtain the depth saliency prediction $P_g$.
The saliency decoder $f_\gamma$ (\enquote{Decoder} in Fig.~\ref{fig:network_overview}) takes the refined saliency features $r_a$, $r_g$, as well as the RGB saliency prediction $P_a$ and depth saliency prediction $P_g$ as input to produce our final prediction $P$, where $\gamma$ is the parameter set of the saliency decoder.
% three parts: the
% % \NB{the} 
% RGB feature $e_a$ and depth feature $e_g$ from the RGB saliency encoder and depth saliency encoder respectively, and the fused saliency after the complementary learning.
% Specifically, the depth saliency from  and depth data and fuses them in a cross-level fusion manner
% % \NB{manner} 
% %NB way
% to produce the fused saliency map $P_f$.
Specifically,
% with the refined RGB saliency feature $r_a$ and refined depth feature $r_g$,
% output $e_a = \{e_a^1,e_a^2,e_a^3,e_a^4\}$ from the RGB saliency encoder and $e_g = \{e_g^1,e_g^2,e_g^3,e_g^4\}$ from the depth saliency encoder, 
we add a position attention module and a channel attention module \cite{fu2019dual} after $r_a$ and $r_g$ to obtain $da(r_a)$ and $da(r_g)$ respectively with discriminative features highlighted.
% each $\{e_a^c\}_{c=1}^4$ and $\{e_g^c\}_{c=1}^4$. 
Then we concatenate $da(r_a)$ and $da(r_g)$, and
% the four groups of feature maps after the dual attention and 
feed it to the DenseASPP~\cite{denseaspp} module to obtain our saliency prediction $P_f$. To further fuse information from both modes, we concatenate $P_a$, $P_g$ and $P_f$ channel-wise, and feed it to a $3\times3$ convolutional layer
%NB s
to achieve our final prediction $P$.

\subsection{Objective function}
\label{obj_fun_sec}
We adopt the binary cross-entropy loss $\mathcal{L}_{ce}$ as our objective function to train our multi-stage cascaded learning framework,
% The variational encoder and the saliency decoder network can be treated as a variational autoencoder (VAE). To train a VAE, one need to maximize the log likelihood of prediction, and at the same time minimize the divergency of the posterior distribution and prior distribution of the latent variable. In our scenario, we define our prediction loss as cross-entropy loss. The latent variable divergency loss is defined as the posterior of $z_r$ or $z_d$ with the prior distribution of the latent variable, which is defined as standard normal distribution $N(0,1)$.
where the complementary constraint, as indicated in Eq.~\eqref{mutual_information}, pushes the saliency feature distribution of the RGB image to be apart from that of the depth data.
Our final objective function is:
\begin{equation}
\label{loss_function}
\begin{aligned}
    \mathcal{L} = & \mathcal{L}_{ce}(P,Y)+\lambda_1\mathcal{L}_{ce}(P_f,Y)+\lambda_2\mathcal{L}_{ce}(P_a,Y)+\\
    &\lambda_3\mathcal{L}_{ce}(P_g,Y)+\lambda \sum_{c=1}^4\mathcal{L}_{mi}(z_a^c,z_g^c),
\end{aligned}
\end{equation}
and empirically we set $\lambda_1=0.8, \lambda_2=0.6, \lambda_3=0.4$.
% \lambda=0.1$ in this paper.
As the range of the $\mathcal{L}_{mi}$ is 10 times larger than that of the $\mathcal{L}_{ce}$, we set its loss weight as $\lambda=0.1$ for balanced learning.

% We introduce mutual information minimization term (MIR), which is used to measure divergency of the distribution of $z_r$ and $z_d$.

% %  LHM \cite{peng2014rgbd}
% %  CDB  \cite{liang2018stereoscopic} 
% %  DESM \cite{cheng2014depth}
% %  GP  \cite{ren2015exploiting}
% %  CDCP \cite{zhu2017innovative}
% %  ACSD \cite{NJU2000}
% %  LBE \cite{feng2016local}
% % DCMC \cite{cong2016saliency}
% % MDSF  \cite{song2017depth}
% % SE  \cite{guo2016salient}
% % DF \cite{qu2017rgbd}$^\dag$
% % AFNet \cite{wang2019adaptive}$^\dag$
% % CTMF \cite{han2017cnns}$^\dag$
% % MMCI \cite{chen2019multi}$^\dag$
% % PCF  \cite{chen2018progressively}$^\dag$
% % TANet \cite{chen2019three}$^\dag$ 
% % CPFP \cite{zhao2019Contrast}$^\dag$
% % D$^3$Net \cite{sip_dataset} 
%  \cite{dmra_iccv19,chen2018progressively,chen2019multi,chen2019three,zhao2019Contrast,select_focus_rgbd,self_attention_rgbd,JLDCF_rgbd,fu2020siamese,a2dele_rgbd,fan2020bbs,zhai2020bifurcated,ji2020accurate,pang2020hierarchical,zhang2020bilateral}. 

\section{\ourdataset~Dataset}
\label{the_new_dataset}
% \NB{you are referring to the existing ones as a set - not sure whether this is plural or not - you maybe want to consider how this is presented - perhaps is an aggregation of three datasets?}
% \NB{This opening paragraph is a bit repeated from your introduction. Do you want to spend space restating?}
% Existing training images for RGB-D saliency detection (the standard composite training dataset) come from two main datasets: (1) NJU2K \cite{NJU2000} and (2) NLPR \cite{peng2014rgbd}. Piao \etal~\cite{dmra_iccv19} introduced an additional 800 images for training and another 400 images for testing to the DUT dataset.
% % which lead to a\NB{n} overall number of 3,000 training image pairs. Meanwhile, 
% As suggested by \cite{dmra_iccv19}, to test model performance on the DUT testing dataset, one needs to train with the
% % dataset existing DUT-RGBD dataset related models are trained with the 
% combination of the NJU2K and NLPR training sets, and fine-tune
% %NB d 
% the model on the DUT training dataset.
% to evaluate on the DUT-RGBD dataset. 
As shown in Table \ref{tab:existing_rgbd_dataset}, the exiting RGB-D saliency detection training dataset is not big enough, which may lead to models of poor generalization ability. Further, as the training dataset is a combination of samples from NJU2K \cite{NJU2000} and NLPR dataset \cite{peng2014rgbd}, different splits
% \NB{s} 
of the training set often lead to inconsistent performance evaluation. Lastly, the small size of testing dataset may fails to full evaluate the RGB-D saliency detection models.
% \cite{peng2014rgbd}  to produce a  compared with the RGB saliency detection counterpart, which may lead to biased model with poor generalization ability.
% % Further, the existing testing dataset may not be  not 
% % We argue that the limited size of the
% % % \NB{size of the} 
% % training set
% % % compared with RGB saliency detection\footnote{The two largest RGB saliency training datasets have 10,553 (DUTS \cite{imagesaliency}) and 10,000 (MSRA10K \cite{msra10k}) images respectively.} 
% % may lead to models with
% % % \NB{s with}
% % %NB of 
% % poor generalization ability.
% Furthermore, different splits
% % \NB{s} 
% of a training set often lead to inconsistent performance evaluation. 
To boost 
%NB the 
RGB-D saliency detection, we contribute the largest RGB-D saliency detection dataset. We provide binary annotation, instance level annotation, ranking based annotation, weak annotation as shown in Fig.~\ref{fig:dataset_annotation_all}. The detailed analysis of the dataset is introduced in the supplementary material.

\subsection{Dataset annotation}
\label{dataset_annotation}
% \YC{One sentence why we resort to stereo images rather than RGB-D images from depth sensors such as Kinect.}
Our new \ourdataset~dataset is based on Holo50K \cite{hua2020holopix50k}, which is a stereo dataset, including scenarios from both indoor and outdoor. We % select
first filter\footnote{We deleted the violent images.} the Holo50K dataset and then obtain
16,000 stereo image pairs for labelling (the candidate labeled set) and another 5,000 image pairs as the unlabeled set. Note that the stereo pairs in Holo50K dataset are directly captured by a stereo camera without rectification, we use a modified version of a SOTA off-the-shelf stereo matching algorithm \cite{zhong2020displacement} to compute the depth for both the candidate labeled set and unlabeled set with the left-right view images as input.
% \Jing{explain the reason why optical flow is the proper to generate depth in our scenario.}

% \YC{Here, we need one sentence on why using the optical flow paper to obtain depth.} 
To provide annotations for the candidate labeled set, we firstly ask five \enquote{coarse} annotators to
% coarsely
% \NB{coursely} 
label salient regions of each image (only the right view image is used) with scribble annotations.
% according to their own preference of saliency.
Secondly, the \enquote{fine} annotators will segment the full scopes of salient objects and provide instance-level annotations.
Thirdly, we perform \enquote{majority voting} to obtain the binary GT saliency maps for our RGB-D saliency detection task. Note that, we delete those samples with no common salient regions, and obtain our final labeled dataset of size 15,625. 
Further, based on the scribble annotations and instance-level saliency maps, we rank each saliency instance according to the initial scribble annotations to form our RGB-D saliency ranking dataset. 

We also provide weak annotations for weakly-supervised RGB-D saliency detection, including scribble annotations and polygon annotations. We define the majority of the scribble annotations from multiple coarse annotators as the scribble annotations of our dataset. Specifically, we first obtain the instance with the majority of scribble. Then, we
% define instance with the majority scribble as the majority instance, and then
define the scribble on the majority instance as
% we obtain instance segmentation of each scribble, and then define scribbles fall on the majority instance as 
our scribble annotation.
% \XY{still not clear to me. instance seg of scribble?}
% define
% \NB{How does majority of scribble work? How do you manage when they sometimes would scarcely overlap - you may have to put this precisely in the supp}
% tWe compute the intersection the all scribbles from the coarse annotators and the binary ground truth saliency map after majority voting to form our scribble annotation. 
We label the majority salient instance with polygons to form our polygon based annotations.

% Saliency ranking dataset:
% For each image, we have instance-level annotation for from multiples annotators. We then sum up those annotation to obtain the ranking based saliency dataset.

% Majority voting:
% Besides the saliency ranking, we can also obtain majority annotations of the multiple instance-level annotation, which can serve as our training dataset for conventional RGB-D saliency detection.

% Weakly supervised saliency detection:
% We can learn from both scribble annotation or polygon annotation. For the former, we have annotation for all the labelers. For the later, we only use the polygon annotation for the majority region.

\begin{table*}[t!]
  \centering
  \scriptsize
  \renewcommand{\arraystretch}{0.9}
  \renewcommand{\tabcolsep}{0.30mm}
  \caption{Benchmarking results of three
  %NB 3 
  leading handcrafted feature-based models and eighteen deep models ($^*$) on six RGB-D saliency datasets.  $\uparrow \& \downarrow$ denote the larger and smaller is better, respectively. Here, we adopt mean $F_{\beta}$ and mean $E_{\xi}$~\cite{Fan2018Enhanced}.}
  \label{tab:BenchmarkResults}
  \begin{tabular}{lr|ccc|cccccc|cccccccccccc|c}
  \hline
%   \toprule
  &  &\multicolumn{3}{c|}{Early Fusion Models}&\multicolumn{6}{c|}{Late Fusion Models}&\multicolumn{12}{c|}{Cross-level Fusion Models}& \\
    & Metric &
   DF & DANet  & UCNet   & LHM  & DESM & CDB &
   A2dele & AFNet & CTMF & JLDCF & DMRA & PCF & MMCI   & TANet   & CPFP & S2MA & BBS-Net & CoNet   & HDFNet& BiaNet & CMWNet & \ourmodel \\
   &  & \cite{qu2017rgbd}$^*$        
   & \cite{DANet}$^*$       
   & \cite{jing2020uc}$^*$      
   & \cite{peng2014rgbd}   
   & \cite{cheng2014depth}                 
   & \cite{liang2018stereoscopic}  
   & \cite{A2dele_cvpr2020}$^*$
   & \cite{wang2019adaptive}$^*$   
   & \cite{han2017cnns}$^*$ 
   & \cite{Fu2020JLDCF}$^*$  
   & \cite{dmra_iccv19}$^*$       
   & \cite{chen2018progressively}$^*$
   & \cite{chen2019multi}$^*$    
   & \cite{chen2019three}$^*$
   & \cite{zhao2019Contrast}$^*$  
   & \cite{self_attention_rgbd}$^*$   
   & \cite{fan2020bbs}$^*$ 
   & \cite{ji2020accurate}$^*$ 
   & \cite{HDFNet-ECCV2020}$^*$ 
   & \cite{zhang2020bilateral}$^*$
%   & \cite{Li_2020_CMWNet}$^*$
   & \cite{cmms_rgbd}$^*$
   & Ours$^*$ \\
  \hline
  \multirow{4}{*}{\rotatebox{90}{\textit{NJU2K}}}%\cite{NJU2000}
    & $S_{\alpha}\uparrow$    & .763 & .897 & .897  & .514 & .665 & .632 & .873 & .822 & .849 & .902 & .886 & .877 & .858 & .879 & .878 & .894 & .921 & .911 & .908 & .915 & .903 & \textbf{.939}  \\
    & $F_{\beta}\uparrow$     & .653 & .877 & .886  & .328 & .550 & .498 & .867 & .827 & .779 & .885 & .873 & .840 & .793 & .841 & .850 & .865 & .902 & .903 & .892 & .903 & .881 & \textbf{.925}   \\
    & $E_{\xi}\uparrow$       & .700 & .926 & .930  & .447 & .590 & .572 & .913 & .867 & .846 & .935 & .920 & .895 & .851 & .895 & .910 & .914 & .938 & .944 & .936 & .934 & .923 & \textbf{.956}  \\
    & $\mathcal{M}\downarrow$ & .140 & .046 & .043  & .205 & .283 & .199 & .051 & .077 & .085 & .041 & .051 & .059 & .079 & .061 & .053 & .053 & .035 & .036 & .038 & .039 & .046 & \textbf{.032}  \\ \hline
    % \midrule
  \multirow{4}{*}{\rotatebox{90}{\textit{SSB}}}%\cite{niu2012leveraging}
    & $S_{\alpha}\uparrow$    & .757 & .892 & .903  & .562 & .642 & .615 & .876 & .825 & .848 & .903 & .835 & .875 & .873 & .871 & .879 & .890 & .908 & .896 & .900 & .904 & .905 & \textbf{.921}   \\
    & $F_{\beta}\uparrow$     & .617 & .857 & .884 & .378 & .519 & .489 & .874 & .806 & .758 & .873 & .837 & .818 & .813 & .828 & .841 & .853 & .883 & .877 & .870 & .879 & .872 & \textbf{.895}   \\
    & $E_{\xi}\uparrow$       & .692 & .915 & .938 & .484 & .579 & .561 & .925 & .872 & .841 & .936 & .879 & .887 & .873 & .893 & .911 & .914 & .928 & .939 & .931 & .926 & .928 & \textbf{.959}  \\
    & $\mathcal{M}\downarrow$ & .141 & .048 & .039 & .172 & .295 & .166 & .044 & .075 & .086 & .040 & .066 & .064 & .068 & .060 & .051 & .051 & .041 & .040 & .041 & .043 & .043 & \textbf{.034}   \\ \hline
    % \midrule
  \multirow{4}{*}{\rotatebox{90}{\textit{DES}}}%\cite{cheng2014depth}
    & $S_{\alpha}\uparrow$    & .752 & .905 & .934 & .578 & .622 & .645 & .881 & .770 & .863 & .931 & .900 & .842 & .848 & .858 & .872 & .941 & .933 & .906 & .926 & .931 & .934 & \textbf{.953}  \\
    & $F_{\beta}\uparrow$     & .604 & .848 & .919 & .345 & .483 & .502 & .868 & .713 & .756 & .907 & .873 & .765 & .735 & .790 & .824 & .909 & .910 & .880 & .910 & .910 & .909 & \textbf{.926}  \\
    & $E_{\xi}\uparrow$       & .684 & .961 & .967 & .477 & .566 & .572 & .913 & .809 & .826 & .959 & .933 & .838 & .825 & .863 & .888 & .952 & .949 & .939 & .957 & .948 & .955 & \textbf{.970}  \\
    & $\mathcal{M}\downarrow$ & .093 & .028 & .019 & .114 & .299 & .100 & .030 & .068 & .055 & .021 & .030 & .049 & .065 & .046 & .038 & .021 & .021 & .026 & .021 & .021 & .022 & \textbf{.015}  \\ \hline
    % \midrule
  \multirow{4}{*}{\rotatebox{90}{\textit{NLPR}}}%\cite{peng2014rgbd}
    & $S_{\alpha}\uparrow$    & .806 & .908 & .920 & .630 & .572 & .632 & .887 & .799 & .860 & .925 & .899 & .874 & .856 & .886 & .888 & .916 & .930 & .900 & .923 & .925 & .917 & \textbf{.941}  \\
    & $F_{\beta}\uparrow$     & .664 & .850 & .891 & .427 & .430 & .421 & .871 & .755 & .740 & .894 & .865 & .802 & .737 & .819 & .840 & .873 & .896 & .859 & .894 & .894 & .877 & \textbf{.909}   \\
    & $E_{\xi}\uparrow$       & .757 & .945 & .951 & .560 & .542 & .567 & .933 & .851 & .840 & .955 & .940 & .887 & .841 & .902 & .918 & .937 & .950 & .937 & .955 & .948 & .939 & \textbf{.964} \\
    & $\mathcal{M}\downarrow$ & .079 & .031 & .025 & .108 & .312 & .108 & .031 & .058 & .056 & .022 & .031 & .044 & .059 & .041 & .036 & .030 & .023 & .030 & .023 & .024 & .029 & \textbf{.019}  \\ \hline
    % \midrule
  \multirow{4}{*}{\rotatebox{90}{\textit{LFSD}}}%\cite{li2014saliency}
    & $S_{\alpha}\uparrow$    & .791 & .845 & .864 & .557 & .722 & .520 & .831 & .738 & .796 & .862 & .847 & .794 & .787 & .801 & .828 & .837 & .864 & .842 & .854 & .845 & .876 & \textbf{.877}  \\
    & $F_{\beta}\uparrow$     & .679 & .826 & .855 & .396 & .612 & .376 & .829 & .736 & .756 & .848 & .845 & .761 & .722 & .771 & .811 & .806 & .843 & .834 & .835 & .834 & .862 & \textbf{.862}  \\
    & $E_{\xi}\uparrow$       & .725 & .872 & .901 & .491 & .638 & .465 & .872 & .796 & .810 & .894 & .893 & .818 & .775 & .821 & .863 & .855 & .883 & .886 & .883 & .871 & .900 & \textbf{.911}   \\
    & $\mathcal{M}\downarrow$ & .138 & .082 & .066 & .211 & .248 & .218 & .076 & .134 & .119 & .070 & .075 & .112 & .132 & .111 & .088 & .094 & .072 & .077 & .077 & .085 & .066 & \textbf{.064}   \\ \hline
    % \midrule
   \multirow{4}{*}{\rotatebox{90}{\textit{SIP}}}% \cite{sip_dataset}
    & $S_{\alpha}\uparrow$    & .653 & .878 & .875 & .511 & .616 & .557 & .826 & .720 & .716 & .880 & .806 & .842 & .833 & .835 & .850 & .872 & .879 & .868 & .886 & .883 & .867 & \textbf{.894}  \\
    & $F_{\beta}\uparrow$     & .465 & .829 & .867 & .287 & .496 & .341 & .827 & .702 & .608 & .873 & .811 & .814 & .771 & .803 & .821 & .854 & .868 & .855 & .875 & .873 & .851 & \textbf{.887}  \\
    & $E_{\xi}\uparrow$       & .565 & .914 & .914 & .437 & .564 & .455 & .887 & .793 & .704 & .918 & .844 & .878 & .845 & .870 & .893 & .905 & .906 & .915 & .923 & .913 & .900 & \textbf{.933}   \\
    & $\mathcal{M}\downarrow$ & .185 & .054 & .051 & .184 & .298 & .192 & .070 & .118 & .139 & .049 & .085 & .071 & .086 & .075 & .064 & .057 & .055 & .054 & .047 & .052 & .062 & \textbf{.044}  \\ \hline
  \end{tabular}
\end{table*}

\subsection{Dataset splitting}
% As shown in Fig. \ref{dataset_annotation}, we provide five different annotations for various RGB-D saliency related tasks. 
% Our new dataset has 20,625 samples, where 15,625 samples are labeled, and the other 5,000 samples are unlabeled.
We divide the labeled set into one training set with 8,025 samples and two different testing sets of size 4,600 and 3,000 respectively, namely the \enquote{Normal} and the \enquote{Difficult} sets. The training dataset is generated by randomly selecting
% \NB{ing} 
8,025 images from the labeled set. For the testing datasets, we intend to introduce two sets of different difficulties. Specifically, we rank the RGB images based on both global and interior contrast, and denote samples with low global contrast and high interior contrast as the difficult samples\footnote{Details about image global and interior contrast are introduced in the supplementary materials.}. Then we have a pool of 1,800 difficult samples $D_d$ and 5,800 normal samples $D_n$. We random select 30\% samples from $D_d$ and 70\% samples from $D_n$ to obtain our \enquote{Normal} testing set,
% and \enquote{Difficult} testing sets of size 4,600 and 3,000 respectively.
% % , 
and the others as \enquote{Difficult} set.

% other as difficult ===============
\section{Experiments}
We compare our method
% the proposed complementary learning framework
\ourmodel~with existing RGB-D saliency detection models,
% \XY{check the name. I think multi-stage cascaded CL? MSC-CLNet?}, 
and report the performance in Table \ref{tab:BenchmarkResults} \& \ref{tab:BenchmarkResults_DUTRGBD}. Furthermore, we retrain the state-of-the-art RGB-D saliency detection models on our new training dataset, and provide the performance of those models on our testing dataset in Table \ref{tab:BenchmarkResults_OurTestSet}.
% We also explore our dataset by providing three benchmark and baseline models on our weak annotations
% % \XY{dataset cannot be weakly supervised. using some weakly supervised baseline models trained on our dataset?} 
% and stereoscopic saliency dataset. 

\begin{table}[t!]
  \centering
  \scriptsize
  \renewcommand{\arraystretch}{1.0}
  \renewcommand{\tabcolsep}{0.7mm}
%   \caption{Benchmarking results of deep RGB-D saliency detection models on DUT-RGBD testing set.}
  \caption{Model performance on DUT~\cite{dmra_iccv19} testing set.}
  \label{tab:BenchmarkResults_DUTRGBD}
  \begin{tabular}{r|cccccccc|c}
  \hline
    Metric & UCNet  & JLDCF &  A2dele & DMRA & CPFP & S2MA & CoNet  & HDFNet & \ourmodel \\    
   & \cite{jing2020uc}          
   & \cite{Fu2020JLDCF}      
   & \cite{A2dele_cvpr2020}
   & \cite{dmra_iccv19}
   & \cite{zhao2019Contrast}  
   & \cite{self_attention_rgbd}  
   & \cite{ji2020accurate} 
   & \cite{HDFNet-ECCV2020} 
   & Ours\\
  \hline
  
%   \multirow{4}{*}{\rotatebox{90}{\textit{DUT-RGBD} \cite{dmra_iccv19}}}
     $S_{\alpha}\uparrow$   & .907 &  .905 & .884 & .886 & .749 & .903 & .919 & .905  & \textbf{.928}  \\
     $F_{\beta}\uparrow$  & .902 &  .884 & .889 & .883 & .695 & .881 & .911 & .889  & \textbf{.921}  \\
    $E_{\xi}\uparrow$     & .931 &  .932 & .924 & .924 & .759 & .926 & .947 & .929  & \textbf{.959}  \\
    $\mathcal{M}\downarrow$  & .038 &  .043 & .043 & .048 & .100 & .044 & .033 & .040  & \textbf{.030}  \\ \hline
  \end{tabular}
\end{table}

\subsection{Setup}
% We introduce complementary learning framework as shown in Fig. \ref{fig:network_overview} for RGB-D saliency detection to explicitly modeling the complementary information from RGB image and depth data.\\
\noindent\textbf{Dataset:} For fair comparisons with existing RGB-D saliency detection models, we follow the conventional training setting, in which the training set is a combination of 1,485 images from the NJU2K
% \fdp{NJN2K$->$NJU2K} 
dataset \cite{NJU2000} and 700 images from the NLPR dataset \cite{peng2014rgbd}. We then test the performance of our model and competing models on the NJU2K testing set, NLPR, testing set 
LFSD \cite{li2014saliency}, DES \cite{cheng2014depth}, SSB \cite{niu2012leveraging} SIP \cite{sip_dataset} and DUT \cite{dmra_iccv19} testing set.

\noindent\textbf{Metrics:} We evaluate the performance of the models on four golden evaluation metrics, \ie, Mean Absolute Error ($\mathcal{M}$), Mean F-measure ($F_{\beta}$), Mean E-measure ($E_{\xi}$)~\cite{Fan2018Enhanced} and S-measure ($S_{\alpha}$)~\cite{fan2017structure}, which are explained in detail in the supplementary materials.

\noindent\textbf{Training details:} Our model is implemented using the \textit{Pytorch} library.
%trained in Pytorch.
% using the ResNet50 \cite{ResHe2015} as backbone as shown in Fig.~\ref{fig:network_overview}. 
The two saliency encoders share the same network structure, and are initialized with ResNet50 \cite{ResHe2015} trained on ImageNet, and the other newly added layers are randomly initialized. We resize all the images and ground truth to the same spatial size of $352\times352$ pixels. We set the maximum epoch as 100, and initial learning rate as 5e-5. We adopt the \enquote{step} learning rate decay policy, and set the decay size as 80 and decay rate as 0.1. The whole training takes 4.5 hours with batch size 5 on an NVIDIA GeForce RTX 2080Ti GPU for the conventional training (NJU2K-train+NLPR-train)
% \fdp{(NJU2K-train+NLPR-train)} 
dataset, and 16 hours with our new training (\ourdataset-train)
% (\fdp{\ourdataset-train}) 
dataset.

\subsection{Model comparison}
\noindent\textbf{Quantitative comparison:} We compare the performance of our \ourmodel~and state-of-the-art RGB-D saliency detection models, and report the performance in Table \ref{tab:BenchmarkResults}. Note that, we use the training set of NJU2K and NLPR as existing deep RGB-D saliency detection models. 
The consistently better performance of our model indicates the effectiveness of our solution.
Further, we observe that the performance gaps of current RGB-D saliency detection are very subtle, \eg, BBS-Net~\cite{fan2020bbs}, CoNet \cite{ji2020accurate}, HDFNet \cite{HDFNet-ECCV2020}, BiaNet \cite{zhang2020bilateral}, and CMWNet~\cite{zhang2020bilateral}, which demonstrates the
% \sout{explains} \NB{demonstrates the} 
necessity of
% \sout{a} 
larger and diverse training and testing datasets for model training and evaluation.

%This implies that current testing datasets may not be sufficiently diverse to clearly distinguish the differences of those models.

\noindent\textbf{Performance on DUT~\cite{dmra_iccv19} dataset:} Some existing RGB-D saliency detection approaches \cite{dmra_iccv19,self_attention_rgbd} fine-tune their models on the DUT training dataset \cite{dmra_iccv19} to evaluate their performance on the DUT testing set. To test our model on the DUT testing set, we follow the same training strategy.
%, and show our performance in Table \ref{tab:BenchmarkResults_DUTRGBD}. 
In Table \ref{tab:BenchmarkResults_DUTRGBD}, all the models are trained with the conventional training set and then fine-tuned on the DUT training set. 
The consistently superior performance illustrates the superiority of our model.
Furthermore, since the current testing performance in Table \ref{tab:BenchmarkResults_DUTRGBD} is achieved in a train-retrain manner (train on the combination training set, and retrain on DUT training set \cite{dmra_iccv19}), we re-train these models with a combination of the conventional training set and DUT training set, and observe consistently worse performance. This observation tells us that inconsistent annotations may occur in the
% \NB{the} 
above three training sets (\ie, NJU2K, NLPR and DUT). It also motivates us to collect a larger training dataset with consistent annotations for robust model training.
% that is training with the conventional training set (combination of NJU2K and NLPR) and fine-tune on the DUT saliency dataset.
% \noindent\textbf{Abaltion study:}

\begin{table*}[t!]
  \centering
  \scriptsize
  \renewcommand{\arraystretch}{0.9}
  \renewcommand{\tabcolsep}{0.55mm}
  \caption{Performance of the extra experiments.
  }
  \begin{tabular}{l|cccc|cccc|cccc|cccc|cccc|cccc}
  \hline
%   \toprule
  &\multicolumn{4}{c|}{NJU2K\cite{NJU2000}}&\multicolumn{4}{c|}{SSB \cite{niu2012leveraging}}&\multicolumn{4}{c|}{DES \cite{cheng2014depth}}&\multicolumn{4}{c|}{NLPR \cite{peng2014rgbd}}&\multicolumn{4}{c|}{LFSD \cite{li2014saliency}}&\multicolumn{4}{c}{SIP \cite{sip_dataset}} \\
    Method 
    & $S_{\alpha}\uparrow$ & $F_{\beta}\uparrow$ & $E_{\xi}\uparrow$ & $\mathcal{M}\downarrow$
    & $S_{\alpha}\uparrow$ & $F_{\beta}\uparrow$ & $E_{\xi}\uparrow$ & $\mathcal{M}\downarrow$
    & $S_{\alpha}\uparrow$ & $F_{\beta}\uparrow$ & $E_{\xi}\uparrow$ & $\mathcal{M}\downarrow$
    & $S_{\alpha}\uparrow$ & $F_{\beta}\uparrow$ & $E_{\xi}\uparrow$ & $\mathcal{M}\downarrow$
    & $S_{\alpha}\uparrow$ & $F_{\beta}\uparrow$ & $E_{\xi}\uparrow$ & $\mathcal{M}\downarrow$
    & $S_{\alpha}\uparrow$ & $F_{\beta}\uparrow$ & $E_{\xi}\uparrow$ & $\mathcal{M}\downarrow$ \\
  \hline
%   &\multicolumn{24}{c}{Model Analysis} \\ \hline
    Base & .910 & .900 & .935 & .035 & .890 & .870 & .917 & .043 & .926 & .915 & .959 & .018 & .920 & .898 & .942 & .024 & .842 & .835 & .880 & .077 & .879 & .876 & .917 & .049    \\
   K3 & .928 & .908 & .947 & \textbf{.032} & .909 & .892 & .939 & .036 & .934 & .922 & .964 & .018 & .925 & .904 & .956 & .022 & .869 & .845 & .898 & .067 & .885 & .879 & .919 & .047    \\
   K32 & .924 & .909 & .944 & .033 & .908 & .894 & .941 & .036 & .938 & .923 & .966 & .017 & .927 & .906 & .959 & .021 & .856 & .853 & .900 & .065 & .885 & .878 & .921 & .046   \\
   SS & .926 & .913 & .943 & .034 & .914 & .882 & .942 & .036 & .946 & .927 & .968 & .017 & .932 & .896 & .954 & .021 & .861 & .852 & .896 & .067 & .885 & .879 & .925 & .046    \\
   W0 & .918 & .907 & .944 & .033 & .892 & .877 & .923 & .042 & .934 & .924 & .964 & .017 & .924 & .900 & .945 & .023 & .843 & .836 & .881 & .076 & .884 & .878 & .916 & .048    \\
   W1 & .919 & .909 & .946 & \textbf{.032} & .905 & .886 & .937 & .037 & .938 & .927 & .971 & .016 & .923 & .903 & .956 & .022 & .857 & .853 & .891 & .071 & .887 & .882 & .921 & .045    \\
    \hline
    $P_f$ & .925 & .908 & .945 & .033 & .908 & .887 & .939 & .036 & .946 & .925 & .965 & .016 & .938 & .907 & .962 & .023 & .862 & .845 & .896 & .068 & .889 & .886 & .927 & .045    \\
%   $z_r$ & . & . & . & . & . & . & . & . & . & . & . & . & . & . & . & . & . & . & . & . & . & . & . & .    \\
   $S_{rgb}$ & .898 & .890 & .930 & .040 & .899 & .876 & .924 & .042 & .891 & .883 & .920 & .028 & .908 & .885 & .932 & .031 & .817 & .807 & .853 & .095 & .860 & .865 & .905 & .056    \\
   $S_{rgbd}$ & .915 & .901 & .932 & .037 & .903 & .878 & .931 & .039 & .920 & .908 & .942 & .021 & .914 & .893 & .943 & .026 & .850 & .841 & .886 & .071 & .876 & .870 & .910 & .051    \\
    \hline
   \ourmodel & \textbf{.939} & \textbf{.925} & \textbf{956} & \textbf{.032} & \textbf{.921} & \textbf{.895} & \textbf{.959}& \textbf{034} & \textbf{.953} & \textbf{.926} & \textbf{.970} & \textbf{.015} & \textbf{.941} & \textbf{.909} & \textbf{.964} & \textbf{.019} & \textbf{.877} & \textbf{.860} & \textbf{.911} & \textbf{.064} & \textbf{.894} & \textbf{.887} & \textbf{.933} & \textbf{.044}    \\
   \hline
  \end{tabular}
  \label{tab:ablation_study_experiments}
%   \vspace{-5mm}
\end{table*}

\noindent\textbf{Qualitative comparison:} We further visualize our prediction in Fig.~\ref{fig:figure1}. The qualitative comparisons demonstrate that with the proposed learning strategy, our model can effectively explore the two modalities for multi-modal learning.
% better complementary information for effective multi-mode learning. 
More results are shown in the supplementary materials.
% \fdp{supplementary materials}.

\noindent\textbf{Model size and running time:} Our model size is 84M, which is comparable with the state-of-the-art models, \eg model size of BBS-Net \cite{fan2020bbs} is 100M.
For inference, our model achieves 10 image/s, which is again comparable with the existing models.

\subsection{Ablation study}
% Three main factors may influence the performance of our model, including: (1) the dimension of the lower-dimensional feature $z_a$ and $z_g$; (2) the structure of the \enquote{Mutual information regularizer} module; and (3) the weight of the mutual information loss $\mathcal{L}_{mi}$. 
We perform the following ablation studies to further analyse the components of our model. We also implement our baseline model without the proposed strategy to highlight the contribution of the mutual information minimization regularizer. Note that, all of these
% \NB{of these} \sout{these}
experiments are trained with the conventional training dataset.

\noindent\textbf{The performance of the baseline model:} To test how our designed encoder and decoder in Fig. \ref{fig:network_overview} performs, we remove the \enquote{Mutual information regularizer} part from our framework\footnote{Detailed structure is shown in the supplementary materials}, and directly concatenate the RGB feature $e_a$ with depth feature $e_g$ and feed it to the decoder. The performance is shown as \enquote{Base} in Table \ref{tab:ablation_study_experiments}. We observe comparable performance of \enquote{Base} compared with existing RGB-D saliency detection models. The inferior performance of \enquote{Base} compared with our final results explains superior performance of the proposed solution of using mutual information as regularizer for redundancy constraining.

\noindent\textbf{The dimension of the feature space:} We set the dimension of lower-dimensional feature embedding ($z_a$ and $z_g$) as $K=6$. To test the impact of feature dimensions on the network performance, we set $K=3$ and $K=32$, and report their performance as \enquote{K3} and \enquote{K32} respectively in Table \ref{tab:ablation_study_experiments}. The experimental results demonstrates that our model achieves relative stable performance with different dimensions of the lower-dimensional feature, while the current dimension with $K=6$ works the best.
% This is because the features from the \enquote{Saliency Encoder} module are representative. 

\noindent\textbf{The structure of the \enquote{Mutual information regularizer} module as shown in Fig.~\ref{fig:network_overview}:} As discussed in Section \ref{latent_feature_sec}, the \enquote{Mutual information regularizer} module is composed of one $3\times3$ convolutional layers
% \XY{i remmeber it was five in methodology} 
and one fully connected layer. One may also achieve this directly from the output of the saliency encoder. Specifically, we can feed the RGB feature and depth feature to two fully connected layers to obtain $z_a$ and $z_g$ respectively. 
In Table \ref{tab:ablation_study_experiments}, we report the performance of our model with this simple setting, marked as \enquote{SS}. 
We observe
% \sout{the} 
performance decreases, which indicates the
% \NB{the} 
necessity of introducing more nonlinearity to effectively extract the feature representation of each mode.
% \NB{e}\sout{al}.

\noindent\textbf{The weight of the mutual information regularizer:} The weight $\lambda$ of the mutual information regularization term controls the level of complementary information. We set $\lambda=0.1$ in this paper to achieve balanced training. We then test how the model performs with different
% smaller or larger ÷
$\lambda$, and set $\lambda=0$ and $\lambda=1$ respectively. We show the performance of those variants in Table \ref{tab:ablation_study_experiments}, denoted by \enquote{W0} and \enquote{W1}. The inferior performance of \enquote{W0} indicates the effectiveness of our complementary information modeling strategy. Further, compared with our performance in Table \ref{tab:BenchmarkResults}, we observe relatively worse performance of \enquote{W1}, which inspires us to further investigate finding an optimal weight for the mutual information regularizer.
% can indeed influence model performance. We will investigate a
% %NB the 
% better strategy to adaptively set the weight of the mutual information in the future.
% Our current setting of the $\lambda$ is   As a multi-mode learning framework, we may need to trade-off between representativeness of each mode (the marginal entropy) and their shared information (the mutual information).
% \XY{this explanation is not convincing. letting a network to learn joint information rather than sparing too much efforts on minimizing mutual information within the allowance of the network capacity.} 
% Therefore, the weight $\lambda$ of the mutual information loss term achieve this trade-off.

% \YC{How about another ablation study, $\lambda =0$, which indicates removing the cross-over constraint in the network training. This will show the importance of complementary learning}

\subsection{Discussion}
\label{model_discussion}
% \noindent\textbf{The learned complementary information:}
\noindent\textbf{The effectiveness of mutual information minimization as regularizer:}
% % By minimizing the mutual information term in Eq. 4, we achieve explicitly complementary information learning. 
% We will visualize the latent feature of each mode
% % and show their mutual exclusion 
% in the revised paper.
% % Meanwhile, our final predictions shown in both Fig. 1 and the supplementary material illustrate that our model can indeed discover the complementary parts of each modal.
We computed the mean absolute cosine similarities of the highest stage feature embeddings ($z_a^4$ and $z_g^4$) for \enquote{W0} from Table~\ref{tab:ablation_study_experiments} (without mutual information minimization as regularizer) and ours,
which are $cosine(z_{a,g}^{M0})=0.90$ and $cosine(z_{a,g}^{Ours})=0.11$
% \fdp{which are 0.90 and 0.11} 
on the NLPR testing dataset. This clearly shows the advantage of our solution in extracting a less correlated feature for each mode. We visualize the learned feature embedding in the supplementary materials.
% Other existing datasets have lower quality depth (\eg, SSB, NLPR, LFSD), in which our method demonstrates its strong advantage in focusing on different aspects of each mode (RGB and depth). However, SIP has high quality depth data,
% %NB Depth of poorer quality is well-covered by existing datasets, such as SSB, NLPR, LFSD, where we show strong advantage (Tab 3). While, depth quality of SIP is good. 
% where each mode shares similar structure information, leading to less complementary parts than the other testing datasets. We can still observe performance improvement for SIP with our complementary learning strategy.

\noindent\textbf{The merging strategy:}
% {~Why not merge $z_a$ and $z_g$ instead of $z$ and $e_{a/g}$?}
In this paper, we produce four different saliency maps as intermediate outputs, including the saliency prediction from both the RGB branch ($P_a$) and depth branch ($P_g$), the feature embedding fusion branch ($P_f$), and our final prediction $P$ by fusing $P_a$, $P_g$ and $P_f$. As $P_f$ has already been
% \NB{been} 
included in
% \NB{in} 
the complementary information of $z_a$ and $z_g$, we define our final prediction as $P_f$ without the final fusion to obtain $P$.
% of $P_a$, $P_g$ and $P_f$.
% \fdp{we define our final prediction as $P_f$ without the final fusion of $P_a$, $P_g$ and $P_f$.} 
The performance is shown in Table~\ref{tab:ablation_study_experiments} \enquote{$P_f$}. We observe 
% To test how our model perform with only $P_f$
% , As shown in Fig. \ref{fig:network_overview}, we merge $z_a$ ($z_g$) with $e_g$ ($e_a$) to maximize the usage of both modes. We can also achieve the merging by simply concatenating  $z_a$ and $z_g$. We keep the saliency prediction from RGB branch $P_a$ and that from the depth branch $P_g$, and only change the 
% Experimentally, we did merge $z_a$ and $z_g$ directly, 
% which led to 
inferior performance of \enquote{$P_f$} compared with our final prediction. The main reason is that $z_a$ and $z_g$ are high-level feature embeddings of the RGB image and depth data, which mainly capture the semantic information. The direct merging of $z_a$ and $z_g$ will generate saliency prediction with less structure accuracy.

\begin{table*}[t!]
  \centering
  \scriptsize
  \renewcommand{\arraystretch}{1.0}
  \renewcommand{\tabcolsep}{0.55mm}
  \caption{Performance of the weakly supervised saliency detection baselines.
  }
  \begin{tabular}{l|cccc|cccc|cccc|cccc|cccc|cccc}
  \hline
%   \toprule
  &\multicolumn{4}{c|}{NJU2K\cite{NJU2000}}&\multicolumn{4}{c|}{SSB\cite{niu2012leveraging}}&\multicolumn{4}{c|}{NLPR~\cite{peng2014rgbd}}&\multicolumn{4}{c|}{SIP~\cite{sip_dataset}}&\multicolumn{4}{c|}{\ourdataset-Normal}&\multicolumn{4}{c}{\ourdataset-Difficult} \\
    Method 
    & $S_{\alpha}\uparrow$ & $F_{\beta}\uparrow$ & $E_{\xi}\uparrow$ & $\mathcal{M}\downarrow$
    & $S_{\alpha}\uparrow$ & $F_{\beta}\uparrow$ & $E_{\xi}\uparrow$ & $\mathcal{M}\downarrow$
    & $S_{\alpha}\uparrow$ & $F_{\beta}\uparrow$ & $E_{\xi}\uparrow$ & $\mathcal{M}\downarrow$
    & $S_{\alpha}\uparrow$ & $F_{\beta}\uparrow$ & $E_{\xi}\uparrow$ & $\mathcal{M}\downarrow$
    & $S_{\alpha}\uparrow$ & $F_{\beta}\uparrow$ & $E_{\xi}\uparrow$ & $\mathcal{M}\downarrow$
    & $S_{\alpha}\uparrow$ & $F_{\beta}\uparrow$ & $E_{\xi}\uparrow$ & $\mathcal{M}\downarrow$ \\
  \hline
%   &\multicolumn{24}{c}{Model Analysis} \\ \hline
   Scribble & .823 & .806 & .869 & .080 & .820 & .803 & .884 & .073 & .820 & .737 & .863 & .058 & .815 & .793 & .888 & .076 & .802 & .780 & .856 & .082 & .767 & .749 & .812 & .115    \\
   Polygon & .847 & .827 & .896 & .065 & .853 & .831 & .913 & .056 & .848 & .789 & .899 & .043 & .846 & .822 & .909 & .060 & .827 & .805 & .884 & .065 & .786 & .774 & .841 & .096  \\
%   M3 & .845 & .829 & .894 & .065 & .850 & .830 & .906 & .057 & .841 & .787 & .896 & .044 & .842 & .819 & .903 & .062 & .825 & .809 & .880 & .066 & .775 & .767 & .827 & .102  \\
   \hline
  \end{tabular}
  \label{tab:weakly_saliency_baseline}
%   \vspace{-5mm}
\end{table*}

\noindent\textbf{The contribution of depth data:} Saliency detection can be achieved merely with the RGB image. As discussed in Section \ref{sec:intro},
% \sout{that}
depth introduces useful geometric information for saliency detection. To verify this conclusion, we train our model (including only the encoder and decoder in Fig.~\ref{fig:network_overview}) with and without depth as input\footnote{The model with depth is an early-fusion model, where depth and RGB image are concatenated at input layer}. The performance is shown in Table \ref{tab:ablation_study_experiments} as \enquote{$S_{rgbd}$} and \enquote{$S_{rgb}$} respectively. The superior performance of \enquote{$S_{rgbd}$} compared with \enquote{$S_{rgb}$} explains contribution of the depth data for saliency detection. We also show examples explaining how depth contributes to saliency detection in the supplementary materials.

\noindent\textbf{Depth generation:} We generate our \ourdataset~dataset
% RGB-D saliency dataset 
with Holo50K~\cite{hua2020holopix50k}, where the stereo pairs are not strictly rectified, which may cause severe matching failures even with state-of-the-art stereo algorithms~\cite{zhong2020nipsstereo}. To solve this issue, we relax the horizontal search in stereo algorithms to both horizontal and vertical search but only keep the horizontal displacements as the stereo disparities. We use a modified stereo matching algorithm~\cite{zhong2020nipsflow} to generate the disparities / depth in our dataset. Further, stereo cameras are widely used in mobile devices, which makes it easier to obtain depth information for both indoor and outdoor.
% , which makes the models trained with our dataset easy to be used in real life.

% Show examples when depth data helps.

% \noindent\textbf{SIP not influenced by different ablation study:}
% Other existing datasets have lower quality depth (\eg, SSB, NLPR, LFSD), in which our method demonstrates its strong advantage in focusing on different aspects of each mode (RGB and depth). However, SIP has high quality depth data,
% %NB Depth of poorer quality is well-covered by existing datasets, such as SSB, NLPR, LFSD, where we show strong advantage (Tab 3). While, depth quality of SIP is good. 
% where each mode shares similar structure information, leading to less complementary parts than the other testing datasets. We can still observe performance improvement for SIP with our complementary learning strategy.

\begin{table}[t!]
  \centering
  \scriptsize
  \renewcommand{\arraystretch}{1.0}
  \renewcommand{\tabcolsep}{0.5mm}
  \caption{Performance on the test sets of our new \ourdataset.}
  \label{tab:BenchmarkResults_OurTestSet}
  \begin{tabular}{lr|cccccccc|c}
  \hline
    & Metric & UCNet  & JLDCF &  A2dele & DMRA & CPFP & S2MA & CoNet  & BBS-Net & \ourmodel\\    
   &  & \cite{jing2020uc}          
   & \cite{Fu2020JLDCF}      
   & \cite{A2dele_cvpr2020}
   & \cite{dmra_iccv19}
   & \cite{zhao2019Contrast}  
   & \cite{self_attention_rgbd}  
   & \cite{ji2020accurate} 
   & \cite{fan2020bbs} 
   & Ours\\
  \hline
  \multirow{4}{*}{\rotatebox{90}{\textit{Normal}}}
     & $S_{\alpha}\uparrow$   & .894 &  .894 & .833 & .782 & .795 & .877 & .820 & .902  & \textbf{.915}  \\
     & $F_{\beta}\uparrow$  & .883 &  .875 & .835 & .744 & .716 & .829 & .796 & .879  & \textbf{.893}  \\
    & $E_{\xi}\uparrow$     & .929 &  .919 & .882 & .812 & .801 & .881 & .850 & .923  & \textbf{.941}  \\
    &  $\mathcal{M}\downarrow$  & .036 &  .042 & .060 & .105 & .104 & .059 & .082 & .039  & \textbf{.033}  \\ \hline
    \multirow{4}{*}{\rotatebox{90}{\textit{Difficult}}}
     & $S_{\alpha}\uparrow$   & .822 &  .845 & .787 & .743 & .770 & .828 & .779 & .853  & \textbf{.867}  \\
     & $F_{\beta}\uparrow$  & .814 &  .832 & .795 & .724 & .704 & .789 & .774 & .834  & \textbf{.852}  \\
    & $E_{\xi}\uparrow$     & .859 &  .870 & .838 & .775 & .776 & .836 & .813 & .876  & \textbf{.893}  \\
    &  $\mathcal{M}\downarrow$  & .079 &  .075 & .092 & .137 & .131 & .092 & .113 & .071  & \textbf{.064}  \\ \hline
  \end{tabular}
\end{table}

\subsection{New benchmarks on \ourdataset}
% Different from existing RGB-D saliency detection dataset, which provides only the binary ground truth saliency maps, making it not straightforward to be used in other related tasks. 
% The straightforward way to use our training dataset is using it for RGB-D saliency detection. 
We provide a new benchmark of state-of-the-art models trained with our new training dataset, and show performance in Table \ref{tab:BenchmarkResults_OurTestSet}. Further, with our rich annotations
% due to the multiple annotations provided in our new dataset 
as shown in Fig.~\ref{fig:dataset_annotation_all}, we discuss another three benchmarks for fully/weakly-supervised learning.
% that would need our annotations.
% We believe that our rich labels 
% can
% %NB would
% motivate future model design.

% \YC{One sentence on the conclusion}
% Our new RGB-D saliency detection dataset provide binary ground truth annotation, instance level annotation, weak annotations and ranking based annotation as shown in Fig. \ref{fig:statistic_dataset}. We further explain how our new dataset can be used to achieve the above mentioned tasks.
% then introduce baseline models for each annotation to boost the RGB-D saliency detection.

\noindent\textbf{Benchmark \#1: Re-train existing RGB-D saliency models with our new training dataset.}
% We introduce a new testing dataset with 7,600 images. 
We split our testing dataset to a moderate-level testing set (\enquote{Normal}) and a hard testing set (\enquote{Difficult}) with 4,600 image pairs and 3,000 pairs respectively.
To test how existing RGB-D saliency detection models perform on our new testing sets, we re-train existing RGB-D saliency detection models with our new training set, and show their performance on the new testing sets in Table \ref{tab:BenchmarkResults_OurTestSet}. The performance gap between those existing techniques illustrates effectiveness of our dataset in both model learning and evaluation.

\noindent\textbf{Benchmark \#2: Stereo saliency detection.}
As our RGB-D saliency dataset is constructed on a stereo dataset \cite{hua2020holopix50k}, we directly train a stereo image pair based saliency object detection model, where the depth is implicitly instead of explicitly obtained from the stereo image pairs.
% instead of RGB-D data, where the depth images are obtained from stereo data.
% obtain depth from the stereo dataset, 
% we can train directly with the stereo image pair. 
Although there exist
%NB s 
some stereoscopic saliency detection models \cite{Stereoscopic_Videos_sal,Stereoscopic_depth_confidence,niu2012leveraging,Saliency_stereo}, all of them take both the RGB image and depth as input.
% The video fixation prediction work in \cite{Intrinsic_stereo_video} introduces an intrinsic depth saliency estimation model for video fixation prediction without explicitly obtaining the depth data. 
In this paper, similar to \cite{Intrinsic_stereo_video}, we design a real\footnote{We define the stereoscopic saliency models taking only the left and right view images as input as the \enquote{real} stereoscopic saliency models.} stereoscopic saliency detection model, and
% As there is no such deep stereoscopic RGB-D saliency detection model\footnote{The existing stereoscopic saliency models take RGB-D as input, we intend to design a pure stereoscopic saliency model which only takes left and right view images as input.} 
% % \XY{give a reference here}
% , 
provide a baseline
% model 
to manifest the potential of our dataset for stereoscopic saliency detection.
%
% There are two existing stereoscopic saliency detection datasets, \ie, the NJU2K testing set \cite{NJU2000} and NJU400 \cite{NJU400} dataset, which include 500 and 400 left-right view image pairs respectively. Together with our new testing sets, we have four stereoscopic saliency detection testing sets as shown in Table~ \ref{tab:stereo_saliency_baseline}. 
We train our stereoscopic saliency detection model on our new training dataset, where the left-right view images are taken as inputs and the GT saliency maps in the right view images are used as supervision. The same encoder and decoder as in Fig.~\ref{fig:network_overview} are adopted for our stereoscopic saliency detection model.
% Instead of explicitly using the depth data,
Specifically, we implicitly model the geometric information by using a cost volume between the saliency encoder for the right and left view images. The performance is shown in~Table \ref{tab:stereo_saliency_baseline}. We explained the architecture and the other stereo saliency datasets in detail in the supplementary material.
% \fdp{supplementary material}.

% \begin{figure}[t!]
%   \begin{center}
%   {\includegraphics[width=0.95\linewidth]{stereo_saliency/stereo_saliency.png}}
%   \end{center}
% %   \vspace{-1mm}
%   \caption{Overview of the proposed stereoscopic saliency detection network.}
%   \label{fig:stereo_saliency_overview} 
% %   \vspace{-4mm}
% \end{figure}

\begin{table}[t!]
  \centering
  \scriptsize
  \renewcommand{\arraystretch}{1.0}
  \renewcommand{\tabcolsep}{0.5mm}
  \caption{Performance of the stereo saliency detection baseline.
  }
  \begin{tabular}{ccc|ccc|ccc|ccc}
  \hline
%   \toprule
  \multicolumn{3}{c|}{NJU2K\cite{NJU2000}}&\multicolumn{3}{c|}{NJU400\cite{NJU400}}&\multicolumn{3}{c|}{\ourdataset-Normal}&\multicolumn{3}{c}{\ourdataset-Difficult} \\
    $S_{\alpha}\uparrow$ & $F_{\beta}\uparrow$ & $\mathcal{M}\downarrow$
    & $S_{\alpha}\uparrow$ & $F_{\beta}\uparrow$ & $\mathcal{M}\downarrow$
    & $S_{\alpha}\uparrow$ & $F_{\beta}\uparrow$ & $\mathcal{M}\downarrow$
    & $S_{\alpha}\uparrow$ & $F_{\beta}\uparrow$ & $\mathcal{M}\downarrow$ \\
  \hline
    .874 & .851 & .056 & .882 & .851 & .044 & .874 & .855 & .047 & .825 & .812 & .080  \\
   \hline
  \end{tabular}
  \label{tab:stereo_saliency_baseline}
%   \vspace{-5mm}
\end{table}

% \noindent\textbf{Weakly-supervised RGB-D saliency detection.}
% % Most of existing RGB-D saliency detection models are fully-supervised, which highly relied on the quality of the training dataset. 
% %While we prepare our fully-annotated training dataset, we generate scribble annotations and polygon annotations as shown in Fig.~\ref{zfig:statistic_dataset} (f) and (g) respectively. 
% %\NB{This second sentence pretty much repeats - merge}
% Inspired by \cite{jing2020weakly}, we also introduce scribble and polygon (see Fig.~\ref{fig:dataset_annotation_all} (f) and (g))
% % \XY{\cf means compare, so i changed here to 'see'.}
% based weakly supervised RGB-D saliency detection networks to further explore our new dataset for weakly supervised learning.

\noindent\textbf{Benchmark \#3 and \#4: Scribble/Polygon as supervision.}
For scribble supervision,
% As no structure information exists in scribble annotation, 
we follow \cite{jing2020weakly} and use the smoothness loss and an auxiliary edge detection branch as a constraint to maintain structure information in the prediction.
% Different from \cite{jing2020weakly}, we use the edge of the prediction as supervision for the edge branch to further reduce the hidden labeling burden.\XY{prediction implies your have trained your network, the labeling burden happens before training. what do you mean hidden labeling?}
% Specifically, as our initial scribble annotations locate inside the salient objects, we then
% % keep the scribble annotations from all the annotators, and 
% create extra background scribble annotations following \cite{jing2020weakly}.
% % \NB{This last sentence seems unclear} 
% In this case, we have both foreground scribbles and background scribbles, and 
We train our scribble supervised RGB-D saliency detection model by concatenating RGB and depth in the input layer, and feed the concatenanted feature to one $3\times3$ convolutional layer to adapt the model in \cite{jing2020weakly}. Performance of the scribble annotation based baseline model is shown in Table \ref{tab:weakly_saliency_baseline} \enquote{Scribble}.
% \noindent\textbf{Benchmark \#4: Polygon as supervision.}
The polygon label is generated after
% while\XY{after majority voting? you can remove the redundant expression here.} we prepare the
majority voting. Fig.~\ref{fig:dataset_annotation_all} (g) shows that the polygon label covers a larger area with better structure information than scribbles.
We
% provide two
% %NB 2 
% baseline model in : (1) 
train directly with polygon annotations as pseudo labels by adopting our model in Fig. \ref{fig:network_overview}, and provide performance of this baseline model in Table \ref{tab:weakly_saliency_baseline} \enquote{Polygon}.

\noindent\textbf{Benchmark analysis:} Our RGB-D saliency benchmark in Table \ref{tab:BenchmarkResults_OurTestSet} shows the superior performance of our method. Furthermore, the gap between state-of-the-art models illustrates the effectiveness of our new testing dataset in model evaluation. Our stereoscopic saliency benchmark in Table \ref{tab:stereo_saliency_baseline} introduces another solution to implicitly use the geometric information. Our two weakly-supervised baselines in Table \ref{tab:weakly_saliency_baseline} provide new options for weakly-supervised RGB-D saliency detection.

\section{Conclusion}
We proposed a multi-stage cascaded learning based RGB-D saliency detection framework that explicitly models complementary information between RGB images and depth data. 
By minimizing the mutual information between these two modalities during training, our model focuses on the diverse parts of each mode rather than the redundant information. 
In this fashion, our model is able to exploit the multi-modal information more effectively.
Further, we introduced the largest RGB-D saliency detection dataset with five types of annotations to prosper the development of fully-/weakly-/un-supervised RGB-D saliency detection tasks.
%supervised RGB-D saliency detection and weakly supervised RGB-D saliency detection. 
Four new benchmarks on 7 datasets and our new dataset demonstrate 
the superiority of our model compared to the existing RGB-D saliency detection techniques.
%
%In addition, as our collected images are captured by stereo cameras, we will collect more RGB-D images photographed by other imaging equipment, such as Kinect, to enrich the diversity of image sources in the future.

% As we obtained stereo images from the Holo50K dataset \cite{hua2020holopix50k}, the computed depth data may not be as good as the one directly captured with depth sensor such as Microsoft Kinect. 
% To further strengthen the new dataset and improve its generalization ability on the diversity of depth from different sources, we will capture another 2,000 images with depth from other source, \eg Microsoft Kinect to form a more robust dataset.

% enlarge the current dataset with data with real depth, \eg Kitti.
% The current depth data is computed from stereo dataset. We intend to 

% Furthermore, our ranking based dataset measures the level of saliency. Inspired by \cite{attention_shift_ranking}, we will capture the sequential shifting of attention for our dataset, both for training and testing, to perform shifting of attention estimation.\XY{too many discussions on datasets. how about remove this?}

{\small
\bibliographystyle{ieee_fullname}
\bibliography{egbib}
}

\end{document}